\documentclass{article}

\usepackage[preprint]{corl_2026}
\usepackage{booktabs}
\usepackage{makecell}
\usepackage{xspace}
\usepackage{graphicx} 
\usepackage{amsmath}
\usepackage{amssymb}
\usepackage{algorithm}
\usepackage{algorithmic}
\usepackage{subcaption}
\usepackage{wrapfig}
\newcommand{\ourmethod}{\textsc{RoboWorld}\xspace}

\newcommand{\ouralgorithm}{\textsc{Step Forcing}\xspace}

\title{RoboWorld: Fast and Reliable Neural Simulators for Generalist Robot Policy Evaluation}

\author{
\makebox[\linewidth][c]{%
Byeongguk Jeon\textsuperscript{*,1,2}
\qquad
Seonghyeon Ye\textsuperscript{*,1}
}\\[0.9em]
\makebox[\linewidth][c]{
JaeHyeok Doo\textsuperscript{1}
\quad
Sungdong Kim\textsuperscript{1,2}
\quad
Minjoon Seo\textsuperscript{1,2}
\quad
Hyungmok Son\textsuperscript{2}
\quad
Kimin Lee\textsuperscript{1,2}
}\\[1.1em]
{\small
\textsuperscript{1}KAIST
\qquad
\textsuperscript{2}Config
\qquad
\textsuperscript{*}Equal contribution.
}\\[0.7em]
{\small
Project Website:
\url{https://byeongguks.github.io/RoboWorld/}
}
}

\begin{document}
\maketitle


\begin{figure*}[h]
\vspace{-3mm}
\centering\includegraphics[width=0.85\linewidth]{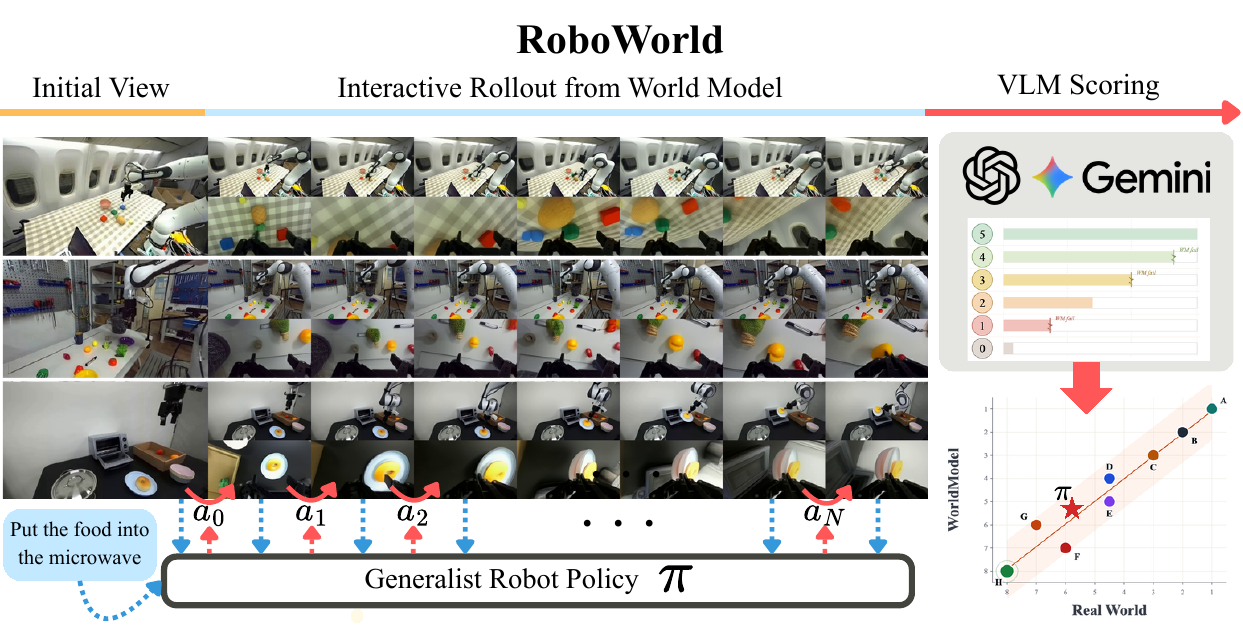}
\caption{
\textbf{Overview of \ourmethod.} \ourmethod evaluates robot policies via closed-loop rollouts in a video world model scored by a task-progress-aware VLM judge, yielding rankings that strongly correlate with real-world evaluations.
}
\label{fig:eval_pipeline}
\vspace{-3mm}
\end{figure*}

\begin{abstract}
  Video world models are emerging as a scalable alternative for evaluating generalist robot policies, bypassing the physical constraints and engineering burdens of real-world deployment.
  However, evaluating policies with video world models remains challenging, as world-model errors can make generated rollouts unreliable and slow inference limits large-scale throughput.
  We introduce \ourmethod{}, an automated evaluation pipeline that pairs a fast autoregressive video world model with a task-progress-aware vision-language model scoring.
  To enable reliable long-horizon autoregressive world-model rollouts, we propose \ouralgorithm{}, which combines anchored and one-step self-forwarded contexts to reduce train--test mismatch while preserving action--observation dynamics.
  Together, these components enable \ourmethod{} to align strongly with real-world robot evaluation across tasks and environments, achieving Pearson's $r = 0.989$ and Spearman's $\rho = 0.970$.

\end{abstract}



\section{Introduction}
\label{sec:intro}

Reliable and accessible evaluation has significantly accelerated progress in foundation models for vision and language~\citep{lin2014microsoft, russakovsky2015imagenet, hendrycks2020measuring, chiang2024chatbot}.
Generalist robot policies, also referred to as Vision-Language-Action (VLA) models, have made rapid progress in generalizing across tasks, objects, and environments, requiring numerous rollouts across diverse conditions for reliable evaluation~\citep{brohan2022rt, zitkovich2023rt, o2024open, team2024octo, kim2024openvla, bjorck2025gr00t, black2024pi0, intelligence2025pi05}. 
However, scaling robot policy evaluation in the real world remains challenging. 
Each rollout requires physical robots and human operators, limiting the number of policies and conditions that can be evaluated.
Simulation-based evaluation~\citep{yu2020meta, james2020rlbench, nasiriany2024robocasa, wang2025roboeval, zhang2025vlabench, kim2026molmospaces} avoids these costs but requires engineering for asset and environment setup, and sim-to-real gaps compromise its reliability~\citep{zhao2020sim, blanco2024benchmarking}.
Recent efforts reduce human intervention~\citep{zhou2025autoeval} or automate real-to-sim environment construction~\citep{jain2025polaris, jangir2025robotarena}, but they still rely on physical robot setups or prior access to target environments.

Built on pretrained video diffusion models~\citep{blattmann2023stable, videoworldsimulators2024, kong2024hunyuanvideo, chen2024videocrafter2, yang2024cogvideox, ali2025world, agarwal2025cosmos, wan2025wan},
video world models offer a scalable alternative for policy evaluation by enabling interactive closed-loop rollouts~\citep{guo2025ctrl, team2025evaluating, gao2026dreamdojo, 1xworldmodel2025, team2026gigabrain, yang2026rise, sharma2026world}.
This property makes it possible to evaluate policies in new environments without physical robot setup or extensive simulator engineering.
However, two challenges remain: (1) world-model artifacts can corrupt long-horizon rollouts, making evaluation unreliable, and (2) slow inference from iterative denoising processes reduces evaluation throughput.
Especially, such artifacts propagate into evaluation outcomes when Vision-Language Model (VLM)-based scoring reduces each rollout to a binary success score~\citep{quevedo2025evaluating, li2025worldeval}.

We introduce \ourmethod, a scalable automated evaluation pipeline that combines a fast autoregressive video world model with a task-progress-aware VLM judge.
To make long-horizon action-conditioned rollouts both fast and reliable, we propose \ouralgorithm{}.
\ouralgorithm{} trains the world model to predict clean frames from one-step self-forwarded priors under the same few-step denoising schedule used at inference.
This alignment reduces the train--test context mismatch and improves few-step denoising quality. Furthermore, interleaving data-grounded anchor contexts ensures the model preserves strict action controllability.
We train our world model on DROID~\citep{khazatsky2024droid} and test its ability to evaluate policies beyond its training environments by measuring the correlation with RoboArena~\citep{atreya2025roboarena}, the largest real-world benchmark built on the DROID setup.
Conditioned on the initial frame of each RoboArena episode, \ourmethod evaluates the same policies inside the world model and produces 4{,}186 video rollouts, achieving a Pearson correlation of $0.989$ and a Spearman correlation of $0.970$ with the original real-world leaderboard.

\begin{figure}[t]
\centering
\includegraphics[width=\linewidth]{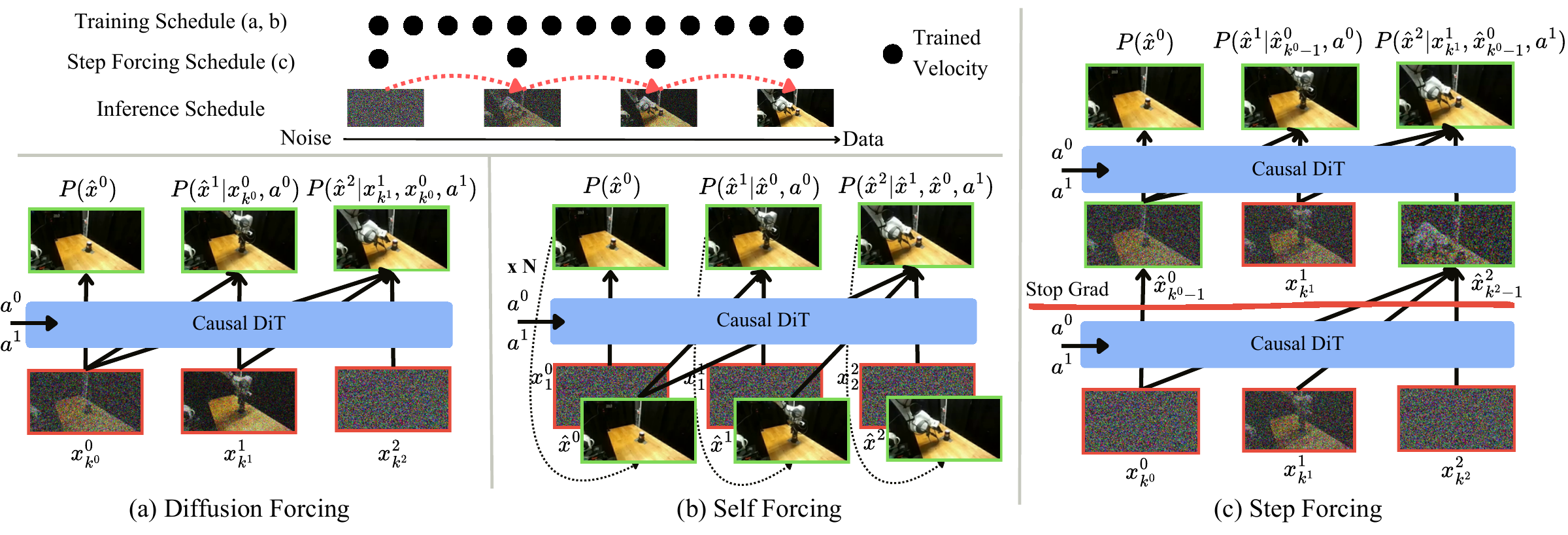}
\caption{\textbf{Left Upper:} \ouralgorithm shares the noise schedule between training and inference.
\textbf{(a)} Diffusion Forcing conditions on noisy ground-truth (red).
\textbf{(b)} Self Forcing conditions on self-generated context (green) via repeated forward rollouts.
\textbf{(c)} \ouralgorithm conditions on the one-step self-forwarded prior (green) or anchor step (red).
}
\vspace{-5mm}
\label{fig:sdf_concept}
\end{figure}

\section{Related Work}
\label{sec:related_work}

\paragraph{Evaluation of Generalist Robot Policies}
Evaluating generalist robot policies~\citep{brohan2022rt, zitkovich2023rt, o2024open, team2024octo, kim2024openvla, bjorck2025gr00t, black2024pi0, intelligence2025pi05} requires comparing policies across diverse tasks, objects, and environments~\citep{nasiriany2024robocasa, atreya2025roboarena}.
Recent real-world benchmarks improve scalability by distributing evaluation or reducing human intervention during reset and success detection~\citep{atreya2025roboarena, zhou2025autoeval}.
Nevertheless, real-world evaluation remains constrained by hardware availability, safety, human oversight, and physical access to evaluation environments~\citep{li2024evaluating}.
Simulation benchmarks provide reproducible rollouts at scale, but require substantial engineering of assets, dynamics, and task environments, and remain limited by sim-to-real gaps~\citep{james2020rlbench, nasiriany2024robocasa, wang2025roboeval, zhang2025vlabench, kim2026molmospaces, tobin2017domain}.
Recent real-to-sim approaches reduce this gap by reconstructing or translating real-world scenes into interactive simulation environments~\citep{jain2025polaris, jangir2025robotarena, zhang2025real, abou2025real}, but still require real-scene access and an explicit simulation-building stage before evaluation.

\paragraph{Video World Models for Robotics}
Video world models have recently been adopted in robotics for modeling future robot behavior across diverse tasks and environments~\citep{gao2026dreamdojo, 1xworldmodel2025, kim2026cosmos, ye2026world, jang2025dreamgen, liao2025genie, li2026causal}.
These models have been used as neural simulators for action-conditioned prediction, planning, and policy rollout~\citep{yang2023learning, wu2024ivideogpt, zhu2024irasim, guo2025ctrl, team2025evaluating, sharma2026world, wang2025learning}.
A closely related line of work explores video world models as proxies for policy evaluation by rolling out robot policies in generative models and automatically scoring the resulting rollouts~\citep{li2025worldeval, tseng2025scalable, quevedo2025evaluating, jiang2025enerverse}.
Despite this progress, slow inference limits evaluation scale, while world-model error can compound over long horizons and bias policy scores.
Also, it remains underexplored how well video-world-model-based evaluation can reflect real-world policy rankings across diverse tasks and environments.

\section{\ouralgorithm}
\label{sec:step_forcing}

In this section, we introduce \ouralgorithm, a training method for fast and reliable action-conditioned long-horizon video generation in video world models.
We begin with the autoregressive video world model (Section~\ref{sec:preliminaries}), then present \ouralgorithm{} (Section~\ref{sec:step_forcing_method}), and finally use BAIR Robot Pushing~\citep{ebert2017self} as a diagnostic setup (Section~\ref{sec:toy-bair}) to motivate our design choices.



\subsection{Preliminaries: Autoregressive video world model}
\label{sec:preliminaries}

Autoregressive video world models~\citep{bruce2024genie, mao2025yume, zhang2025matrix, he2025matrix, valevski2024diffusion} factorize future observations $x^{1:N}$ given an initial observation $x^{0}$ and action sequence $a^{1:N}$ as $p_\theta(x^{1:N} \mid x^0, a^{1:N}) = \prod_{i=1}^{N} p_\theta(x^i \mid x^{0:i-1}, a^i)$, enabling generation beyond the training horizon and efficient inference with
key--value caching~\citep{yin2025slow}.
However, this leads to a train--test context mismatch, as the model is trained with clean or noised ground-truth contexts $x^{<i}$, whereas inference conditions on self-generated predictions $\hat{x}^{<i}$, causing errors to accumulate over long horizons.
Training on self-generated contexts~\citep{huang2025self, guo2025end} mitigates this context mismatch, but incurs substantial overhead from sequential autoregressive rollouts. 
Moreover, replacing $x^{<i}$ with $\hat{x}^{<i}$ shifts the state on which action $a^i$ acts; supervising $p_\theta(x^i \mid \hat{x}^{<i}, a^i)$ thus introduces transitions inconsistent with the action-observation dynamics in the data.

\subsection{\ouralgorithm: Mitigating Train-Test Gap through Self-Forwarded Prior Denoising}
\label{sec:step_forcing_method}

\begin{algorithm}[tb]
\caption{\ouralgorithm}
\label{alg:self_df}
\begin{algorithmic}[1]
\REQUIRE Dataset $\mathcal{D}$ of (video, action) pairs
\REQUIRE Denoising schedule $\{t_0,\dots,t_S\}$ with $0=t_0<\dots<t_S=1$, where $S$ denotes the number of denoising steps.
\REQUIRE Velocity network $v_\theta$, number of frames $N$, anchor probability $p$
\REPEAT
    \STATE Sample $(x, a)\sim\mathcal{D}$
    \STATE Sample $k^i\sim\mathcal{U}\{1,\dots,S\}$, $\epsilon^i\sim\mathcal{N}(0,I)$ 
           for $i\in\{1,\dots,N\}$
    \STATE $x^i_{k^i} \leftarrow t_{k^i}\,\epsilon^i + (1-t_{k^i})\,x^i$ 
           \COMMENT{noisy frames}
    \STATE $h^i \leftarrow \{x^j_{k^j}\}_{j<i}$
           \COMMENT{context frames}
    \STATE With probability $p$, set $\Delta t^i \leftarrow 0$ \emph{(anchor)}; 
       otherwise $\Delta t^i \leftarrow t_{k^i}-t_{k^i-1}$ \emph{(self-forward)}
    \STATE $\tilde{t}^i \leftarrow t_{k^i} - \Delta t^i$ 
       \COMMENT{actual noise level of prior}
    \STATE $\hat{x}^i_{k^i-1} \leftarrow x^i_{k^i} - \Delta t^i \cdot 
           \mathrm{sg}\!\left[v_\theta(x^i_{k^i}, t_{k^i}, h^i, a^i)\right]$ 
           \COMMENT{one-step self-forwarded prior}
    \STATE $\hat{h}^i \leftarrow \{\hat{x}^j_{k^j-1}\}_{j<i}$
           \COMMENT{self-forwarded context frames}
    \STATE $\hat{x}^i \leftarrow \hat{x}^i_{k^i-1} - \tilde{t}^i\,
           v_\theta(\hat{x}^i_{k^i-1}, \tilde{t}^i, \hat{h}^i, a^i)$ 
           \COMMENT{clean prediction}
    \STATE Update $\theta$ by descending 
           $\nabla_\theta \tfrac{1}{N}\sum_{i=1}^{N}\|x^i - \hat{x}^i\|_2^2$
\UNTIL{converged}
\end{algorithmic}
\end{algorithm}

We propose \ouralgorithm, training the world model to denoise from one-step self-forwarded priors under a fixed discrete denoising schedule.
We formulate denoising with rectified flow~\citep{lipman2022flow, liu2022flow}, where a clean frame $x^i$ and noise $\epsilon^i$ are connected by the linear path $x^i_t = t\epsilon^i + (1-t)x^i$.
We define the inference noise schedule $0=t_0<\cdots<t_S=1$ and use the same schedule for one-step self-forwarding during training.
This schedule alignment reduces the train--test mismatch and enables efficient few-step denoising (e.g., $S=4$) without relying on distillation-based few-step samplers~\citep{yin2025slow, huang2025self}.

Formally, given a video-action trajectory $(x^{0:N}, a^{1:N})\sim\mathcal{D}$ and rectified-flow velocity $v_\theta$~\citep{liu2022flow}, we form per-frame noisy latents $x^i_{k^i}$ from independently sampled noise levels $k^i\sim\mathcal{U}\{1,\dots,S\}$, following Diffusion Forcing~\citep{chen2024diffusion}. 
Given the noisy context $h^i := \{x^j_{k^j}\}_{j<i}$, we take a single Euler step under stop gradient to obtain a \emph{self-forwarded prior}:
\begin{equation}\label{eq:self_forward}
\hat{x}^i_{k^i-1} 
= x^i_{k^i} - (t_{k^i}-t_{k^i-1})\,
  \mathrm{sg}\!\left[v_\theta(x^i_{k^i}, t_{k^i}, h^i, a^i)\right],
\end{equation}
where the stop-gradient, $\mathrm{sg[\cdot]}$, prevents $v_\theta$ from being optimized to shape its own future inputs.
One-step self-forwarding mitigates the train-test gap by training the model to learn from its own imperfect generation. 
However, unlike self-generated contexts, it enables \textit{parallel} context generation, significantly reducing the training cost and enabling stable training. 
Also, to ground supervision in the data distribution, with probability $p$ we apply an \emph{anchor step} that bypasses self-forwarding and sets the prior directly to the noisy ground-truth latent, $\hat{x}^i_{k^i-1} = x^i_{k^i}$.
By preventing action-observation mismatch from self-generated context, anchor step preserves ground-truth dynamics, which is essential for action-conditioned video generation.
We then predict the clean frame from $\hat{x}^i_{k^i-1}$ conditioned on the self-forwarded context $\hat{h}^i := \{\hat{x}^j_{k^j-1}\}_{j<i}$, and minimize the squared error to $x^i$; the overall procedure is illustrated in Figure~\ref{fig:sdf_concept}, while the full equations are provided in Algorithm~\ref{alg:self_df}.

\begin{wraptable}{r}{0.45\linewidth}
  \centering
  \vspace{-15pt}


  \centering
  \small
  \setlength{\tabcolsep}{3pt}
  \label{tab:bair_main}

  \resizebox{\linewidth}{!}{%
  \begin{tabular}{lc|cc|cc}
  \toprule
  Method & Steps & \multicolumn{2}{c|}{SSIM $\uparrow$} &
  \multicolumn{2}{c}{LPIPS $\downarrow$} \\
   &  & ID & OOD & ID & OOD \\
  \midrule
  Teacher Forcing~\citep{zhou2025taming}      & 8 & 0.7942 & 0.7118 & 0.0554 & 0.1058 \\
  Diffusion Forcing~\citep{chen2024diffusion} & 8 & 0.7657 & 0.6861 & 0.0690 & 0.1117 \\
  Resampling Forcing~\citep{guo2025end}       & 8 & 0.7891 & 0.6996 & 0.0572 & 0.1082 \\
  Self Forcing~\citep{huang2025self}               & 4 & 0.7929 & 0.6953 & 0.0548 & 0.1083 \\
  \ouralgorithm                                    & 4 & \textbf{0.8063} & \textbf{0.7374} & \textbf{0.0525} & \textbf{0.0768} \\
  \bottomrule
  \end{tabular}
  }


  \caption{Diagnostic comparison of action-conditioned training on BAIR Robot Pushing.}
  \label{tab:bair}
    \centering
  \vspace{-2pt}
  \includegraphics[width=\linewidth]{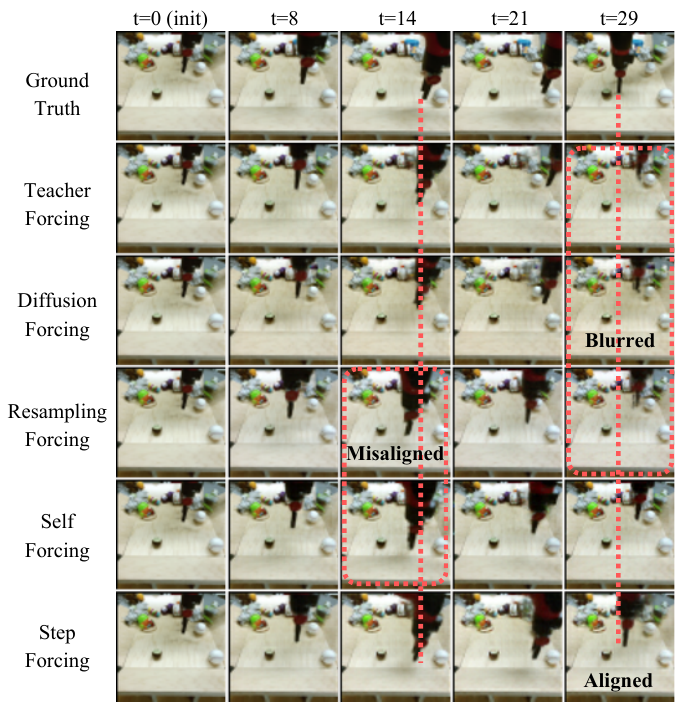}
  \captionof{figure}{\ouralgorithm shows strong action controllability while maintaining visual quality throughout the whole horizon.}
  \label{fig:bair_poc}
  \vspace{-20pt}
\end{wraptable}

\subsection{Diagnostic Comparison on BAIR Robot Pushing}
\label{sec:toy-bair}




We use BAIR Robot Pushing~\citep{ebert2017self} as a small-scale diagnostic setup for comparing different training objectives in action-conditioned autoregressive world modeling. 
We examine whether each objective preserves action--observation dynamics while maintaining stable long-horizon rollouts. Specifically, we measure both in-horizon (ID) and extrapolation (OOD) frames. Experimental details are provided in Appendix~\ref{app:toy-bair}. 



\ouralgorithm achieves the best SSIM~\citep{wang2004ssim} and LPIPS~\citep{zhang2018unreasonable} in both windows (Table~\ref{tab:bair}; Figure~\ref{fig:bair_poc}). 
More importantly, the baselines expose the failure modes motivating our design. 
Teacher Forcing preserves data-grounded transitions but suffers from error accumulation during long-horizon autoregressive rollout. 
Diffusion Forcing improves robustness to noisy contexts, but its generated rollouts can still drift over extrapolation horizons. 
Resampling Forcing and Self Forcing produce visually plausible videos by training on self-generated contexts, but this weakens action following because the same action is supervised from model-induced states rather than data-grounded states. 
In contrast, \ouralgorithm reduces the train--test context gap through one-step self-forwarded priors while using anchor steps to preserve action--observation dynamics. 

\section{\ourmethod}
\label{sec:worldarena}

We present \ourmethod, a fully automated evaluation pipeline for generalist robot policies that scales across diverse environments without requiring physical rollouts or human intervention.
\ourmethod consists of two components: (1) a fast autoregressive video world model trained with \ouralgorithm, and (2) a task-progress-aware VLM judge that scores the generated rollouts.


\subsection{Adapting Pretrained Video Models into Fast Autoregressive World Models}

To adapt pretrained bidirectional video models into fast autoregressive video world models, we make three architectural and training modifications:
(1) we replace bidirectional attention with frame-level causal attention, 
(2) we encode actions with a two-layer MLP and inject them into each frame through cross-attention, and 
(3) we use per-frame independent noise scheduling during training~\citep{chen2024diffusion}.
We first train the model with rectified flow matching~\citep{lipman2022flow} under the Diffusion Forcing scheme~\citep{chen2024diffusion}, converting the bidirectional video model into an autoregressive video world model.
We then apply \ouralgorithm{} to enable stable long-horizon autoregressive rollouts with few-step inference.
During inference, we evaluate the policy in a closed loop within the world model. At each step, the policy predicts an action given the current observation, and the world model generates the next observation conditioned on the past context and the predicted action. This generated observation is then fed back to the policy to predict the subsequent action. To accelerate this autoregressive decoding, following prior video diffusion models~\citep{gao2024ca2, cui2025self, huang2025self, yin2025slow}, we use key--value (KV) caching with a sliding window context.
\subsection{Automatically Scoring Rollouts from World Models}
\label{sec:mitigate_world_model_err}
After the world model generates action-conditioned rollouts, a scoring model is needed to evaluate the policy performance conditioned on the rollout. Previous works rely on a VLM that generates a binary score given the generated rollout \citep{quevedo2025evaluating, li2025worldeval}. However, a binary success metric conflates policy failures with world model artifacts. For example, when a policy picks up the correct object but fails to place it due to world model error (object disappearance), a binary scorer would evaluate the trajectory as a failure.
Instead, we introduce a task-progress-aware evaluation using a predefined 0–5 rubric prompted to a VLM judge. To reliably disentangle true policy failures from world-model artifacts, we instruct the VLM to isolate world-model error detection to the wrist view—where physical inconsistencies predominantly manifest—while assessing actual task progress strictly through the fixed external views. This multi-view evaluation strategy allows us to allocate partial scores reflecting the policy's valid progress before the world-model corruption. Unlike binary metrics, this approach fairly credits policy capabilities under imperfect world-model rollouts. We provide the full VLM prompts and rubric definitions in Appendix~\ref{app:vlm-details}.

\section{Experiments}

We investigate whether \ourmethod can serve as a substitute for real-world evaluations of generalist robot policies.
We train our video world model on DROID~\citep{khazatsky2024droid}, a large-scale real-world robot manipulation dataset, and evaluate on RoboArena \citep{atreya2025roboarena} trajectories.
First, we compare our world model trained with \ouralgorithm against other video world model baselines on long-horizon RoboArena trajectories.
We then measure the correlation between \ourmethod and real-world evaluations by applying \ourmethod to the initial observations from RoboArena evaluation episodes.\footnote{
We evaluate the 8 policies that were open-sourced as of the latest RoboArena data dump used in our experiments (Feb.\ 3, 2026).
}

\subsection{Experimental Details}

We initialize the video world model from Wan2.1-T2V-1.3B~\citep{wan2025wan}.
Following prior multi-view video generation setups~\citep{jang2025dreamgen, ye2026world}, we concatenate DROID's three camera views (two fixed views and one wrist-mounted view) into a $2\times2$ grid and treat it as a single multi-view frame.
We condition the model on end-effector Cartesian position~\citep{huang2025vid2world, guo2025ctrl}; for joint-velocity policies, we use a shallow action adapter to map policy actions into this conditioning space~\citep{guo2025ctrl}.
We train for 160k Diffusion Forcing steps followed by 40k \ouralgorithm steps, using 45-frame clips, a learning rate of $1\times10^{-5}$, batch size $8$, 1k warm-up steps.
At inference time, we use a sliding context window of $W=45$ for open-loop evaluation (Section~\ref{sec:openloop}) and $W=77$ for closed-loop evaluation (Section~\ref{sec:closedloop}).

\begin{figure}[t]
  \centering
  \includegraphics[width=\linewidth]{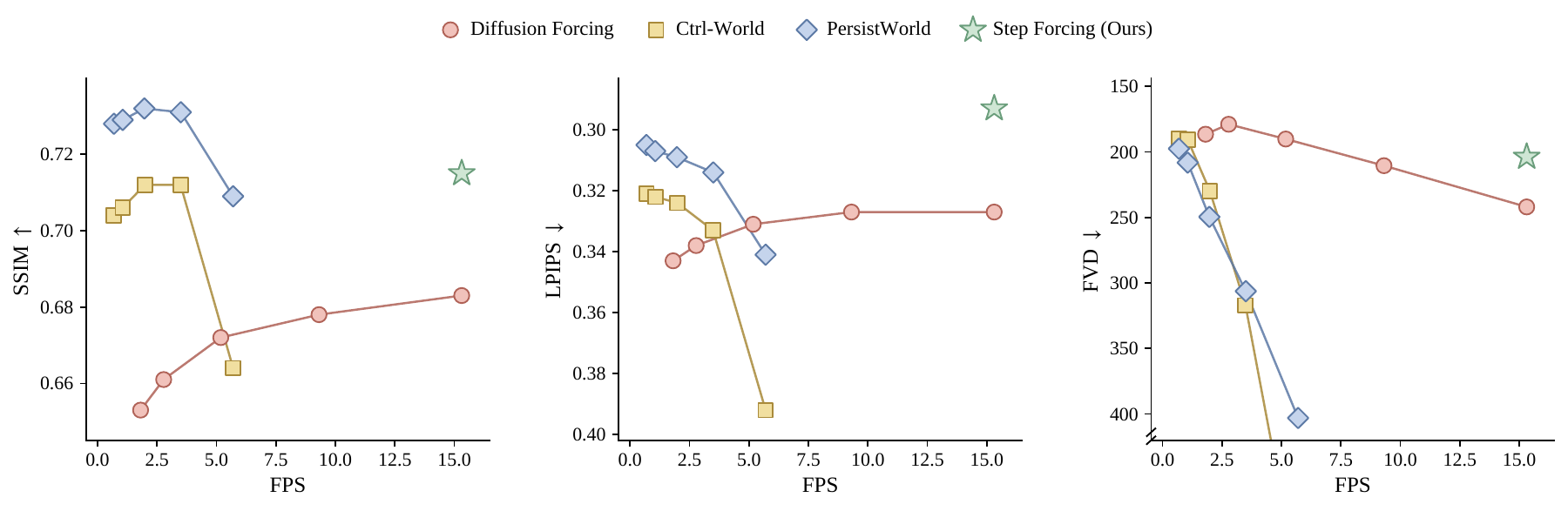}
  \caption{Long-horizon action-conditioned video generation on RoboArena. Quality (SSIM, LPIPS, FVD) vs. FPS. Baseline markers sweep denoising steps $\in$ \{50, 32, 16, 8, 4\}. 
  \ouralgorithm achieves the best LPIPS at the highest FPS (15.3), and matches the baselines that run substantially slower on SSIM and FVD.}
  \vspace{-5mm}
  \label{fig:metrics_vs_fps_combined}
\end{figure}

\subsection{Efficient Long-Horizon Action-Conditioned Video Generation}
\label{sec:openloop}
We evaluate the long-horizon generation quality and inference efficiency of our world model.
We compare against prior video world models~\citep{guo2025ctrl, bardhan2026persistent}, as well as an autoregressive video world model (Diffusion Forcing) baseline trained for the same total number of steps.
For multi-view long-horizon evaluation, we sample $256$ policy-rollout trajectories from held-out RoboArena~\citep{atreya2025roboarena}.
For each trajectory, we condition the world model on the initial multi-view observation and action sequences to generate a $300$-frame ($20$ seconds) video.
We evaluate the generated videos using pixel-level metrics (SSIM~\citep{wang2004ssim}) and perceptual/distributional metrics (LPIPS~\citep{zhang2018unreasonable}, FVD~\citep{unterthiner2018towards}).
All metrics are computed per view, and we report the average over the three views.

Figure~\ref{fig:metrics_vs_fps_combined} compares \ouralgorithm with video world model baselines (Ctrl-World~\citep{guo2025ctrl} and PersistWorld~\citep{bardhan2026persistent}) and Diffusion Forcing baseline in terms of inference throughput (FPS) and generation quality (SSIM, LPIPS, FVD).
For the baselines, we sweep the number of inference denoising steps over $\{50, 32, 16, 8, 4\}$.
Because \ouralgorithm uses the same denoising schedule during training and inference (Section~\ref{sec:step_forcing}), we evaluate it at its training schedule of $4$ steps.
Ctrl-World and PersistWorld use bidirectional attention, preventing KV-cache reuse across frames; at $4$ steps, they run at 5.70 FPS, compared with 15.31 FPS for autoregressive variants.
\ouralgorithm achieves the best LPIPS overall and the best SSIM and FVD among $4$-step methods, providing a substantially better speed--quality tradeoff for long-horizon evaluation compared to baseline models.

\begin{figure}[t]
  \centering
  \includegraphics[width=0.95\textwidth]{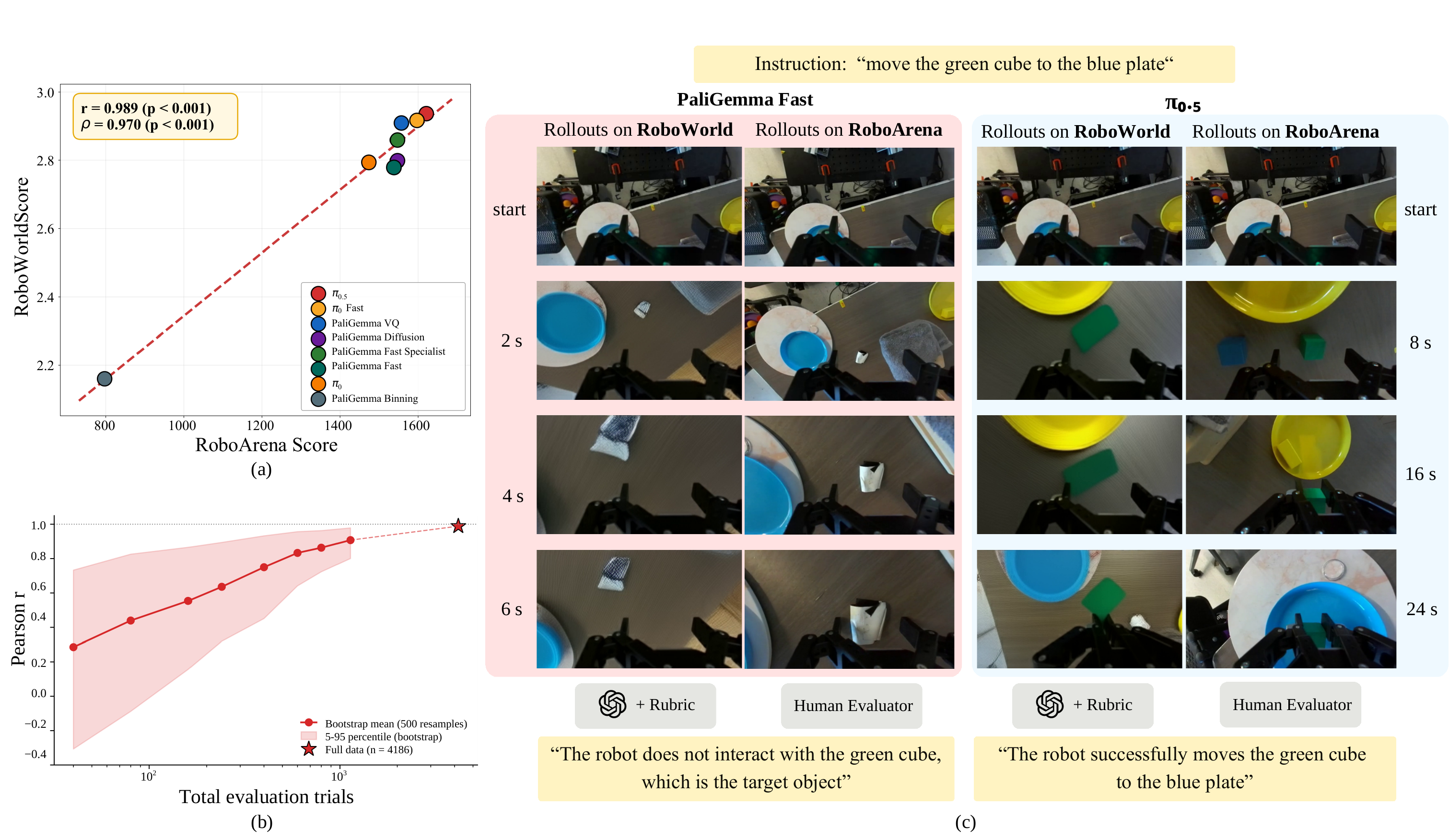}
  \caption{
(a) Pearson and Spearman correlations between \ourmethod and the RoboArena leaderboard.
(b) Pearson correlation under varying numbers of evaluation trials, showing that greater diversity in environments and tasks improves the correlation.
(c) \ourmethod closed-loop rollouts compared with RoboArena rollouts using the same policies and tasks, sharing only the initial view.
Rollouts proceed from top to bottom: PaliGemma Fast fails in both settings by selecting the wrong object, whereas $\pi_{0.5}$ succeeds in both.
}
\vspace{-5mm}
  \label{fig:correlation_plot}
\end{figure}

\subsection{Correlation with Real World Evaluation}
\label{sec:closedloop}
To investigate whether \ourmethod provides reliable evaluation results, we simulate RoboArena evaluation episodes within our world model and measure the correlation with real-world evaluation.
For each policy, we reuse the initial observations from RoboArena episodes that contain all three views: two fixed external views and one wrist view.\footnote{We include all three-view initial observations available in the latest RoboArena data dump.}
Starting from these initial observations, the policy is executed in closed loop within the world model: at each step, the policy predicts an action from the current generated observation, and the world model predicts the next observation conditioned on that action.
This replaces the physical environment in the RoboArena evaluation protocol with our world model while keeping the same policies and initial conditions.
Each rollout proceeds for $30$ seconds and is scored by a VLM judge.
We use GPT-4o as the default judge.\footnote{We use the RoboArena leaderboard snapshot dated Feb.\ 26, 2026}


\ourmethod shows strong positive correlation with the RoboArena leaderboard across the eight evaluated policies, achieving Pearson $r=0.989$ and Spearman $\rho=0.970$ (Figure~\ref{fig:correlation_plot}a).
The resulting policy ranking closely matches the real-world leaderboard under GPT-4o scoring.
Replicating the RoboArena benchmark in our neural simulator (across 8 different policies) takes only 100 H100 GPU hours, including the computation and communication costs for each policy action generation and world-model rollout, substantially reducing the cost of large-scale policy evaluation.
We further analyze how the number of evaluation trials affects correlation with the real-world leaderboard (Figure~\ref{fig:correlation_plot}b).
The correlation improves as the number of trials increases, highlighting the importance of broad coverage for reliable policy comparison.
Qualitative examples in Figure~\ref{fig:correlation_plot}c show that \ourmethod captures both successful executions and policy-specific failure modes.
For example, \ourmethod models the successful behavior of $\pi_{0.5}$ while also reproducing the wrong-object selection failure observed for PaliGemma Fast in RoboArena.
Additional implementation details, qualitative examples, and failure-case analyses are provided in Appendix~\ref{app:closed_loop_qualitative}.

\subsection{Ablations and Analyses}

\begin{wrapfigure}{r}{0.42\linewidth}


    \centering
    \vspace{-10pt}
    
    \includegraphics[width=\linewidth]{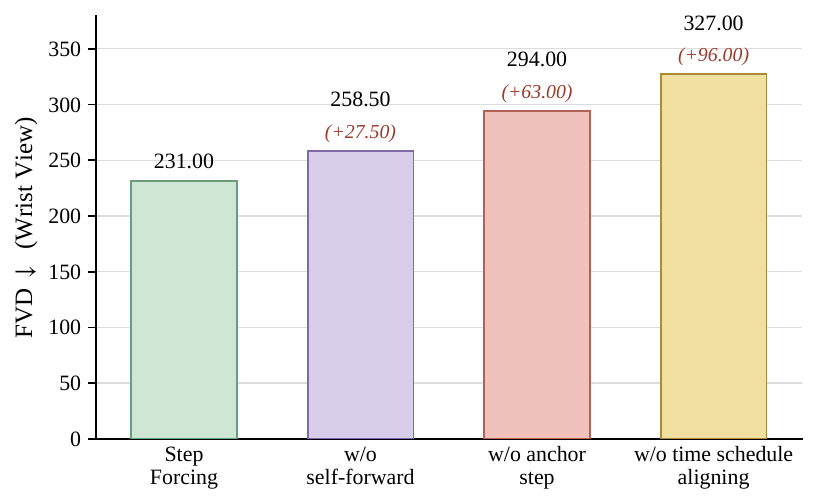}
    \caption{Component ablation of \ouralgorithm{} on DROID.}
    \vspace{-5mm}
    \label{fig:ablation_fvd}
    
    \vspace{-12pt}






\end{wrapfigure}

\paragraph{Component ablation of \ouralgorithm{}.}
Figure~\ref{fig:ablation_fvd} ablates the main components of \ouralgorithm{} on DROID, mainly focusing on the wrist view, the most challenging view due to its fast and action-sensitive motion.
All results are evaluated with 4-step denoising during autoregressive rollout.
The full method achieves an FVD of $231.0$, while independently removing the self-forwarded prior and anchor step increases FVD to $258.5$ and $294.0$, respectively.
These results indicate that schedule alignment is crucial for stable 4-step autoregressive rollout, and that both self-forwarding and anchor steps substantially reduce drift in the wrist view.
We also observe consistent findings on BAIR Robot Pushing setup; see  Table~\ref{tab:bair_main_full} in Appendix~\ref{app:toy-bair}.

\paragraph{Effect of the VLM Evaluation Rubric.}
\label{sec:ablation-vlm}
We ablate the task-progress rubric (§\ref{sec:mitigate_world_model_err}) by replacing \ourmethod{} scores with binary scores. 
We measure Spearman correlation between scores and the RoboArena leaderboard ranking across eight policies. The task-progress rubric achieves $\rho = 0.970$, while binary success rate reduces this to $\rho = 0.922$.
The rubric's sensitivity to partial task progress is important for reliable policy ranking.

\begin{figure}[t]
  \centering
  \includegraphics[width=\linewidth]{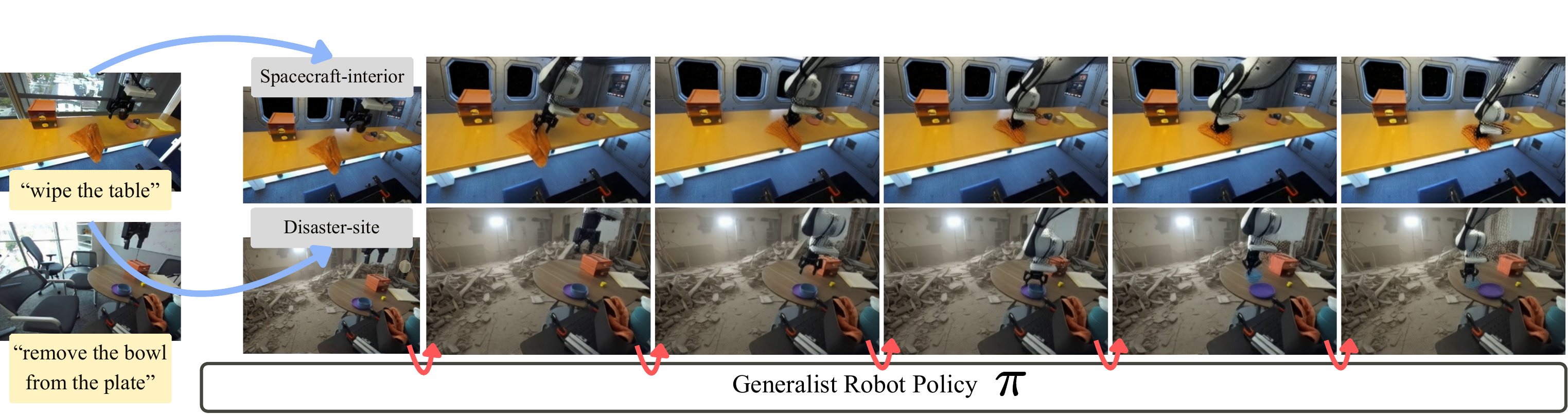}
  \caption{
    Qualitative examples of \ourmethod{} in synthetic extreme environments. 
    We transform real-world robot images into extreme environment scenes (e.g., spacecraft interiors, disaster sites), and conduct closed-loop policy evaluation.
    }
\vspace{-5mm}
  \label{fig:synthetic_rollout}
\end{figure}

\paragraph{Extending \ourmethod to Extreme Environments}
We investigate whether \ourmethod could be applied to extreme environments (i.e., construction sites or airplane cabins), where real-world robot evaluation at scale is infeasible. From 175 initial observations from RoboArena snapshots, we generate eight synthetic environments (airplane cabin, spacecraft interior, operating room, nuclear facility, underwater station, disaster site, construction site, mine tunnel) utilizing an image editing model and retain 746 valid initial conditions after manual filtering. As shown in Figure \ref{fig:synthetic_rollout}, \ourmethod can be robustly applied in extreme environments without physical access or complex asset generation for simulation. Moreover, the correlation between \ourmethod on extreme environments and RoboArena leaderboard is still mainly retained ($r=0.970$), implying that \ourmethod could be utilized for reliable and safe policy evaluation before physical deployment to these environments.

\section{Conclusion}

We present \ourmethod, a scalable evaluation pipeline for generalist robot policies built on a fast autoregressive video world model and a task-progress-aware VLM judge.
To enable reliable long-horizon rollouts, we train the world model with \ouralgorithm, which reconstructs clean frames from self-forwarded priors and data-derived anchor contexts.
Across RoboArena policies, \ourmethod achieves high correlation with real-world evaluation, showing that world-model rollouts can approximate real-world policy rankings while substantially reducing evaluation cost.

\section{Limitations}
Despite rapid progress, video world models for dexterous, long-horizon robot manipulation evaluation remain limited. A key challenge is that long-horizon manipulation requires maintaining object consistency under sustained contact. Achieving this typically necessitates substantial training data that capture rich embodied interactions with objects. A promising direction to alleviate this limitation is to pretrain on large-scale human--object interaction data, which provides broad coverage of object behaviors and contact dynamics.

\clearpage


\bibliography{example}  

\newpage
\appendix
\begin{center}{\bf {\LARGE Appendix}}
\end{center}

\section{Implementation Details: Action-conditioned BAIR}
\label{app:toy-bair}

\paragraph{Dataset.}
We use the BAIR Robot Pushing dataset~\citep{ebert2017self}: $43{,}264$ training clips and $256$ test clips of $30$ frames each, $64{\times}64$ RGB pixels in $[-1, 1]$, accompanied by a per-frame $4$-dimensional robot action. 
Training samples a random $15$-frame window per clip; evaluation takes the first $30$ frames of each test clip and uses $K{=}1$ as the conditioning prefix. All experiments use the original $64{\times}64$ resolution.

\paragraph{Architecture Details.}
We use a causal video DiT~\citep{peebles2023scalable} adapted to the $3$-channel RGB pixel space (Table~\ref{tab:bair_arch}). 
Each frame is patchified by a $4{\times}4$ Conv2d into $16{\times}16{=}256$ spatial tokens. 
Each DiT block contains spatial self-attention (per frame), causal temporal self-attention (across frames), and an MLP, with all three components AdaLN-modulated by a per-frame conditioning vector.
Temporal position is encoded with Rotary Position Embeddings
(RoPE)~\citep{su2024roformer} applied to the temporal-attention
$Q,K$ projections. 
The action vector is projected through a two-layer MLP and added to the per-frame condition that drives AdaLN.

\begin{table}[h]
  \centering
  \caption{Backbone hyperparameters.}
  \label{tab:bair_arch}
  \begin{tabular}{lcc}
    \toprule
    & student / main & teacher \\
    \midrule
    embed\_dim       & 256 & 448 \\
    depth            & 6   & 12  \\
    num\_heads       & 8   & 14  \\
    patch\_size      & 4   & 4   \\
    RoPE max\_seq    & 128 & 128 \\
    attention        & causal & bidirectional \\
    action MLP       & $4, 512, 256$ & $4, 896, 448$ \\
    parameter count  & $9.2$M & $54.6$M \\
    \bottomrule
  \end{tabular}
\end{table}
\paragraph{Loss Function.}
All five methods share the same backbone and the same loss target ($v$-prediction).
\begin{itemize}
\item \textbf{Diffusion Forcing}~\citep{chen2024diffusion}: per-frame
independent noise levels $t_i \sim \mathcal{U}(0.001, 0.999)$. A
single forward pass through the causal DiT predicts $v_i$ for every frame and the loss is averaged.
\item \textbf{Teacher Forcing} ~\citep{zhou2025taming}: per-frame independent $t_i \sim \mathcal{U}(0.001, 0.999)$ as in DF, but each target frame $n_i$ should be predicted conditional on \emph{strict-past clean} frames $c_{<i}$, not on the (noisy) future. Naively this is one forward pass per target $i$ ($T$ passes per step). Following Resampling Forcing~\citep{guo2025end}, we use the 2T-sequence dense trick: pack inputs as $[c_0, ..., c_{T-1}, n_0, ..., n_{T-1}]$, share temporal indices between $c_i$ and $n_i$, and apply the block-attention mask shown below. This delivers one supervised prediction per frame in a single forward.
\item \textbf{Resampling Forcing}~\citep{guo2025end}: Because the official code is not publicly available, we implemented faithfully according to the paper. Each training step has two phases: (i) \emph{Resampling}: sample a shared timestep $t_s\sim\text{LogitNormal}(\text{shift}{=}0.6)$, corrupt all frames at $t_s$ with i.i.d.\ Gaussian noise, then \emph{sequentially} 1-step Euler-denoise per frame in \texttt{no\_grad} to obtain resampled clean estimates $\tilde{x}_{0..T-1}$; (ii) \emph{Supervised step}: sample per-frame independent $t_i\sim\mathcal{U}(0.001,0.999)$, pack $[\tilde{x}_0,\dots,\tilde{x}_{T-1},\,n_0,\dots,n_{T-1}]$ with shared RoPE indices, and apply the same 2T-dense mask as in TF.
\item \textbf{Self Forcing}~\citep{huang2025self}: implemented based on the official code, with the following modifications motivated by our small-scale BAIR setup:
\begin{enumerate}
      \item \emph{initial view conditioning.}
      For a fair comparison, we use the same 9M-parameter causal DiT backbone as the other BAIR baselines, whereas the original Self Forcing implementation is built on Wan-1.3B-i2v. Accordingly, the conditioning interface differs slightly: the original implementation uses channel-axis image-to-video conditioning, while our BAIR setup uses frame-axis conditioning. Specifically, the ground-truth conditioning frame is concatenated along the temporal axis and pinned at $t{=}0$. To keep the score networks in-distribution during distillation, we prepend this conditioning frame at $t{=}0$ to every \texttt{real\_score} and \texttt{fake\_score} call.

      \item \emph{Numerical stabilization.}
      For numerical stability, we clamp the DMD2 normalizer $\|x_{0,\text{real}}-x_{0,\text{student}}\|_1$ from below by a small constant $\varepsilon{=}0.1$. This avoids division-by-zero when the teacher nearly perfectly reconstructs the student output.
\end{enumerate}
All other hyperparameters follow the paper: $\textsc{Lr}_G{=}2\!\times\!10^{-6}$, $\textsc{Lr}_F{=}4\!\times\!10^{-7}$, AdamW $\beta_1{=}0,\beta_2{=}0.999$, weight\_decay${=}0.01$, EMA${=}0.99$ from step $200$, critic:generator update ratio $5{:}1$.
\item \textbf{\ouralgorithm}: trained at a fixed discrete schedule $S{=}4$, with per-frame timestep sampled from $\{0.25, 0.50, 0.75, 1.0\}$. We supervise the model on self-forwarded denoising trajectories with anchor probability $p{=}0.5$, as described in Section~\ref{sec:step_forcing}.
\end{itemize}

\paragraph{2T-sequence dense mask (TF).}
Let $T$ be the training-window length. We concatenate $T$ clean and
$T$ noisy frames along the temporal axis and share RoPE positions
$[0,\dots,T-1, 0,\dots,T-1]$. The boolean attention mask
$M\in\{0,1\}^{2T\times 2T}$ is:
\begin{align*}
M_{q,kv}\!=\!1\;\;\text{iff}\;\;
\begin{cases}
q<T,\; kv\!\leq\! q                  & \text{(clean half causal)},\\
q{=}T{+}i,\; kv<i                    & \text{(noisy $n_i$ sees strict-past clean)},\\
q{=}T{+}i,\; kv{=}q                  & \text{(noisy self-attention)}.
\end{cases}
\end{align*}
The mask blocks noisy-to-noisy attention, ensuring that each noisy
target is predicted without conditioning on other noisy targets under
the dense layout. This yields the same per-target conditioning
structure as the naive $T$-pass TF objective, up to a constant factor
in the loss, while requiring only a single forward pass per step.

\paragraph{Training.}
We use AdamW with a constant learning rate of $2\!\times\!10^{-4}$,
$\beta=(0.9,0.999)$, weight decay $10^{-4}$, and global gradient-norm
clipping at $1.0$. We train for $100$k steps with batch size $16$.

\paragraph{Training Cost Comparison.}
We compare the per-step training cost of each method in our implementation.
All models have 9.2M parameters and are trained on a single NVIDIA H100.
``Fwd/step'' denotes the number of model forward passes required per
optimizer step, including no-gradient forwards used for autoregressive
resampling.

\begin{center}
\begin{tabular}{lccc}
\toprule
Method & Fwd/step & sec/step & step/s \\
\midrule
Diffusion Forcing                  & 1                    & 86 ms          & 11.6 \\
\ouralgorithm                      & 2                    & 122 ms         & 8.2  \\
Teacher Forcing                    & 1                    & 180 ms         & 5.5  \\
Resampling Forcing                 & $T{+}1$ ($16$)       & $\sim\!1.3$ s  & 0.77 \\
Self Forcing~\citep{huang2025self} & $\sim\!344$          & 2.4 s          & 0.42 \\
\bottomrule
\end{tabular}
\end{center}

\footnote{\emph{sec/step} is wall-clock time measured on a single H100;
\emph{Fwd/step} is an analytical count of model forward passes per optimizer step,
including no-gradient forwards. DF uses one gradient forward. \ouralgorithm uses
two forwards: one no-gradient self-forward and one gradient forward. TF uses one
forward over the $2T$-packed sequence. RF uses $T{+}1$ forwards: $T$ no-gradient
resampling forwards plus one supervised forward. Since our training window length
is $T{=}15$, this gives $16$ forwards per optimizer step. For SF (DMD), with a
$5{:}1$ critic-to-generator update ratio, each optimizer step uses six
autoregressive rollouts of $F\!\cdot\!S$ generator forwards plus score/critic
evaluations, giving $6\cdot56 + 3 + 5 = 344$ at $F{=}14$ and $S{=}4$.}

\noindent
Teacher Forcing uses a $2T$-length dense sequence, which increases
temporal-attention cost relative to Diffusion Forcing.
\ouralgorithm uses two forward passes per training step: one no-gradient
self-forward pass and one gradient update pass, both over a length-$T$ sequence.
Resampling Forcing performs an autoregressive resampling loop with $T$ sequential
no-gradient forwards, one per target frame, before the supervised update.
For Self Forcing, we follow the original DMD distillation procedure, which
requires multi-step autoregressive rollouts together with repeated score and
critic evaluations; this leads to a higher per-step cost in the BAIR setting.

\medskip\noindent
\textbf{Implementation note.}
The wall-clock numbers above reflect our straightforward implementation of each method.
In particular, our autoregressive-style baselines use dense causal-mask forwards without key--value cache amortization.
Thus, the reported costs should be interpreted as a practical comparison under a common implementation rather than as the most optimized possible runtime for each algorithm.
Further engineering, such as caching key--value tensors across autoregressive resampling steps, could reduce the absolute cost of RF- or SF-style methods.
However, these methods still require sequential autoregressive resampling during training, whereas \ouralgorithm only adds a single no-gradient self-forward pass.
Therefore, we expect the qualitative cost ordering to remain similar:
\[
\text{DF} < \ouralgorithm < \text{TF} \ll \text{RF} \ll \text{SF}.
\]
Overall, this comparison highlights that autoregressive resampling-based training objectives can be substantially more expensive than objectives that avoid long sequential resampling loops.

\paragraph{Inference.}
At evaluation time, we autoregressively generate $30$ frames starting from $K{=}1$ clean conditioning frame and the ground-truth action sequence. 
For simplify the implementation we do not use KV-caching.

\paragraph{Evaluation protocol.}
We evaluate on $64$ held-out test clips (BAIR test split, deterministic
seed). Each rollout is compared frame-by-frame against the ground-truth
clip with:
\begin{itemize}
  \item \textbf{MSE}: pixel-space mean squared error on $[-1,1]$.
  \item \textbf{SSIM}~\citep{wang2004ssim}: window size $11$,
    \texttt{pytorch\_msssim} default; computed on rescaled $[0,1]$ frames.
  \item \textbf{LPIPS}~\citep{zhang2018unreasonable}: AlexNet backbone,
    computed directly on $[-1,1]$ frames.
\end{itemize}

\paragraph{Full Results.}
Table~\ref{tab:bair_main_full} extends the main paper table with oracle teacher checkpoints ($9$M / $55$M) and the FVD column. The teachers are evaluated under their own training protocol, using in-painting at $T{=}15$ and $S{=}16$. Therefore, they are not directly comparable to the causal student rows, but serve as an oracle ceiling reference.

\begin{table}[h]
\centering
\small
\setlength{\tabcolsep}{3pt}
\caption{Full BAIR high-motion (top-64) evaluation. Causal students autoregressively roll out $29$ frames from one conditioning frame at $S{=}4$, while teachers use in-painting at $T{=}15$ and $S{=}16$. Frame~$0$ is excluded from ID metrics.}
\label{tab:bair_main_full}

\resizebox{\linewidth}{!}{%
\begin{tabular}{lcc|ccc|ccc|c}
\toprule
Method & Fwd/step & $S$ & \multicolumn{3}{c|}{In-distribution (1--14)} &
\multicolumn{3}{c|}{Out-of-distribution (15--29)} & FVD $\downarrow$ \\
       &          &     & MSE $\downarrow$ & SSIM $\uparrow$ & LPIPS $\downarrow$ &
                         MSE $\downarrow$ & SSIM $\uparrow$ & LPIPS $\downarrow$ & \\
\midrule
\multicolumn{10}{l}{\emph{Teachers (oracle reference)}} \\
Teacher 9M  & 1 & 16 & 0.0658 & 0.7820 & 0.0583 & --- & --- & --- & 20.42 \\
Teacher 55M & 1 & 16 & \textbf{0.0587} & \textbf{0.8069} & \textbf{0.0489} & --- & --- & --- & \textbf{14.46} \\
\midrule
\multicolumn{10}{l}{\emph{Causal students}} \\


Teacher Forcing~\citep{zhou2025taming}      & 1                    & 8 & 0.0620 & 0.7942 & 0.0554 & 0.1185 & 0.7118 & 0.1058 & \textbf{17.70} \\
Diffusion Forcing~\citep{chen2024diffusion} & 1                    & 8 & 0.0758 & 0.7657 & 0.0690 & 0.1113 & 0.6861 & 0.1117 & 22.47 \\
Resampling Forcing~\citep{guo2025end}       & $T{+}1$              & 8 & 0.0618 & 0.7891 & 0.0572 & 0.1265 & 0.6996 & 0.1082 & 19.15 \\
Self Forcing~\citep{huang2025self}          & $\sim\!344^{\dagger}$ & 4 & 0.0588 & 0.7929 & 0.0548 & 0.1246 & 0.6953 & 0.1083 & 21.31 \\
\textbf{Step Forcing (ours)}                & \textbf{2}           & 4 & \textbf{0.0512} & \textbf{0.8063} & \textbf{0.0525} & \textbf{0.0778} & \textbf{0.7374} & \textbf{0.0768} & 19.26 \\
\midrule
\multicolumn{10}{l}{\emph{Ablations of Step Forcing}} \\
$-$\,continuous-$t$              & 2 & 4  & 0.0660 & 0.7895 & 0.0670 & 0.0922 & 0.7318 & 0.0959 & 18.43 \\
$-$\,continuous-$t$              & 2 & 8  & 0.0700 & 0.7859 & 0.0651 & 0.0966 & 0.7274 & 0.0952 & 18.03 \\
$-$\,continuous-$t$              & 2 & 16 & 0.0720 & 0.7834 & 0.0640 & 0.0998 & 0.7226 & 0.0951 & 18.31 \\
$-$\,no self-forward ($p{=}1$)   & 1 & 4  & 0.1329 & 0.7001 & 0.1081 & 0.1587 & 0.6298 & 0.1401 & 29.91 \\
$-$\,no anchor ($p{=}0$)         & 2 & 4  & 0.1869 & 0.5998 & 0.1933 & 0.4141 & 0.3745 & 0.3334 & 49.25 \\
$-$\,DF (discrete)               & 1 & 4  & 0.0911 & 0.7002 & 0.0913 & 0.1583 & 0.5797 & 0.1451 & 43.63 \\
\bottomrule
\end{tabular}}

\smallskip
{\footnotesize
$^{\dagger}$Self Forcing's $\sim\!344$ forwards per training step decomposes as one generator update ($56$ autoregressive rollout forwards $+\,3$ score evaluations) plus five critic updates, each consisting of $56$ autoregressive rollout forwards $+\,1$ score evaluation. This gives $59 + 5\times57 = 344$.}
\end{table}

\section{Vision-Language Model-Based Evaluation Details}
\label{app:vlm-details}
This section provides additional details on the VLM-based evaluation procedure used in \ourmethod. We describe the motivation behind using a VLM-as-a-judge, the scoring rubric designed to separate policy progress from world-model artifacts, the multi-view evaluation strategy, the rollout sampling and scoring protocol, the exact prompt templates, and ablations validating these design choices.

\subsection{VLM-as-a-Judge: Rubric and Strategy}
\label{app:vlm_rubric}

\begin{figure}[h]
\centering
\includegraphics[width=0.7\linewidth]{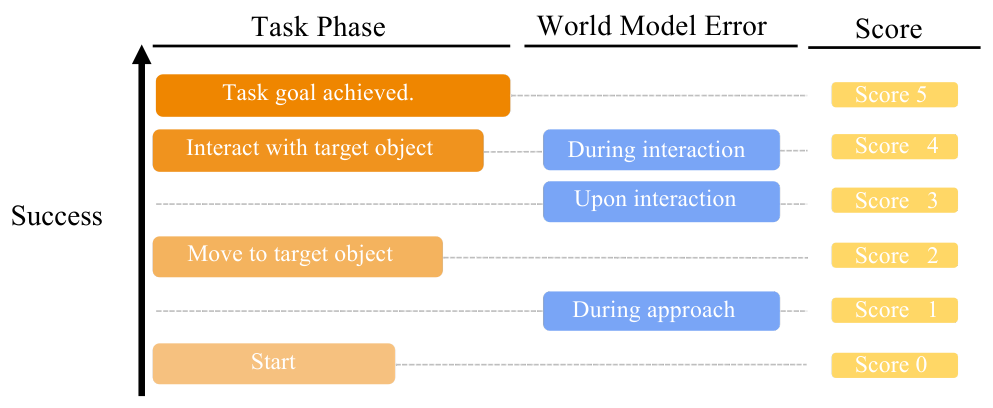}
\caption{\textbf{Design principles of the evaluation rubric.} We 
design the rubric such that scores primarily reflect task progress. 
In addition, we assign different penalties depending on when 
world-model errors occur, so failures that arise earlier in the 
rollout receive lower scores than those that happen after 
substantial task progress.}
\label{fig:vlm_rubric}
\end{figure}

Human scoring requires human-in-the-loop evaluation, limiting 
scalability. To maximize scalability, we replace human annotators 
with a pretrained VLM as a judge, which generates a score when 
conditioned on rollouts produced by the video world model. However, 
since world model rollouts inevitably contain model-induced 
artifacts, naive automatic scoring may compromise reliability. To 
maintain reliable evaluation, we introduce two design choices.

\paragraph{(1) Fine-grained evaluation rubric.}
We design a six-level rubric (Figure~\ref{fig:vlm_rubric}) that 
assigns scores based on both task progress and the stage at which 
world-model errors occur. Concretely:
\begin{itemize}
\setlength\itemsep{0pt}
\item \textbf{5 (Success)}: Task goal is achieved.
\item \textbf{4 (Near success, world-model fail during interaction)}: 
The robot reaches and interacts with the target.
\item \textbf{3 (World-model fail upon interaction)}: World-model 
failure after target contact.
\item \textbf{2 (Attempt without interaction)}: The robot approaches 
the target object.
\item \textbf{1 (World-model fail before interaction)}: World-model 
failure while approaching the target.
\item \textbf{0 (Failure)}: All other cases.
\end{itemize}

\paragraph{(2) Anchoring success judgments on stable fixed views.}
To ensure reliable evaluation, we primarily anchor task success judgments on the two fixed external views, which provide a stable and consistent perspective of the workspace. In contrast, the wrist view is highly dynamic and prone to localized world-model artifacts due to rapid camera motion and contact dynamics. Therefore, we utilize the fixed views as the definitive signal for assessing task progress and completion, while restricting the wrist view's role solely to the identification of world-model failures.
\subsection{VLM Evaluation Procedure}
\label{app:vlm_eval_procedure}

For VLM-based evaluation, we use GPT-4o as the judge. Each rollout 
is sampled at one frame per second and split into two segments. We 
score the segments independently using GPT-4o, providing 15 sampled 
frames per segment along with the rubric and the task description. 
The final per-rollout score is the maximum across the two segments. 
To reduce evaluation cost, we omit scoring the second segment when 
the policy is already judged successful in the first segment.

\subsection{VLM Prompt Templates}
\label{app:vlm_prompts}

Table~\ref{tab:vlm_rubric} presents the rubric used to compute the 
score. Tables~\ref{tab:vlm_prompt_with_wrist} 
and~\ref{tab:vlm_prompt_without_wrist} present the prompts used for 
the vision-language model evaluator.

\begin{table*}[t]
\centering
\small
\setlength{\tabcolsep}{8pt}
\renewcommand{\arraystretch}{1.15}
\begin{tabular}{p{0.95\textwidth}}
\toprule
\textbf{Evaluation Rubric Used in the VLM Prompt} \\
\midrule

\textbf{Evaluation Criteria}

\vspace{0.3em}
\textbf{Target object:} The object specified in the task instruction.

\vspace{0.5em}
\textbf{Score 5: Success}\\
\textbf{Definition:} The policy successfully completes the task as instructed.\\
$\bullet$ The target object is correctly manipulated to reach the intended goal state.\\
$\bullet$ The outcome is clearly successful and verifiable.\\
$\bullet$ Success should not be judged from a single frame.\\
$\bullet$ Evaluators must check adjacent frames to confirm that the goal state is genuinely achieved and stably maintained.\\
To determine success, the robot must be clearly observed interacting with the target object.

\vspace{0.5em}
\textbf{Score 4: Near success OR World Model failure during target-object contact/interaction}\\
\textbf{Definition:} The policy nearly completes the task but does not fully succeed, OR a world model failure occurs after the policy has made contact with (or begun manipulating) the target object.\\
$\bullet$ The policy correctly reaches and interacts with the target object, but the final goal is not achieved.

\vspace{0.5em}
\textbf{Score 3: World Model failure upon target-object contact/interaction}\\
\textbf{Definition:} A world model failure occurs at the moment the policy begins interacting with the target object.\\
$\bullet$ The world model fails immediately when the robot makes contact with the target object or begins manipulation.

\vspace{0.5em}
\textbf{Score 2: Attempted execution near target}\\
\textbf{Definition:} The policy attempts to solve the task and moves near the target object.\\
$\bullet$ The robot moves near the target object with task-directed behavior.

\vspace{0.5em}
\textbf{Score 1: World Model failure during approach before target-object interaction}\\
\textbf{Definition:} A world model failure occurs while moving toward the target object.\\
$\bullet$ The robot is near the target object in the fixed view, but the target object disappears or is not visible in the wrist view.\\
$\bullet$ Evaluation is unreliable due to this fixed-view vs.\ wrist-view inconsistency.

\vspace{0.5em}
\textbf{Score 0: Failure (All other cases)}\\
\textbf{Definition:} Any case not covered by Scores 5, 4, 3, 2, or 1.\\
$\bullet$ The policy exhibits movements irrelevant to the task instruction.\\
$\bullet$ The robot moves independently of the target object in the fixed view.\\
$\bullet$ The robot leaves the scene in the fixed view.\\
$\bullet$ World model failures that occur while the robot is not meaningfully interacting with or approaching the target object. \\

\bottomrule
\end{tabular}
\caption{Evaluation rubric used in the VLM prompt.}
\label{tab:vlm_rubric}
\end{table*}

\begin{table*}[t]
\centering
\small
\setlength{\tabcolsep}{8pt}
\renewcommand{\arraystretch}{1.15}
\begin{tabular}{p{0.95\textwidth}}
\toprule
\textbf{Prompt Template Used for VLM Evaluation (Wrist View Used Only for Error Identification)} \\
\midrule

You are evaluating a robot policy rollout video. The frames below are extracted from the video in chronological order (first frame to last frame).

\vspace{0.3em}
\textbf{Task Instruction:} \texttt{\{instruction\}}

\vspace{0.3em}
\texttt{\{rubric\}}

\vspace{0.3em}
Multiview frames are provided, corresponding to a robot policy rollout in the world model.  
Each frame consists of a fixed left view (upper right view), a fixed right view (upper left view), and a wrist view (bottom left).

\vspace{0.5em}
\textbf{Evaluation Method:}

\vspace{0.3em}
1. Use the fixed views (the two upper views) as the primary reference.\\
Determine whether the robot interacts with the target object based on the fixed views.

\vspace{0.3em}
2. Use the wrist view only to identify world model errors (e.g., object disappearance or inconsistency).\\
Do NOT use the wrist view to infer task success.

\vspace{0.5em}
\textbf{Provide:}

\vspace{0.3em}
1. A score (0 to \texttt{\{max\_score\}}) according to the rubric above\\
2. A brief explanation of your reasoning \\

\bottomrule
\end{tabular}
\caption{Prompt template used when the wrist view is used only to identify world model errors, but not to determine task success.}
\label{tab:vlm_prompt_without_wrist}
\end{table*}

\begin{table*}[t]
\centering
\small
\setlength{\tabcolsep}{8pt}
\renewcommand{\arraystretch}{1.15}
\begin{tabular}{p{0.95\textwidth}}
\toprule
\textbf{Prompt Template Used for VLM Evaluation (Wrist View Used for Full Evaluation)} \\
\midrule

You are evaluating a robot policy rollout video. The frames below are extracted from the video in chronological order (first frame to last frame).

\vspace{0.3em}
\textbf{Task Instruction:} \texttt{\{instruction\}}

\vspace{0.3em}
\texttt{\{rubric\}}

\vspace{0.3em}
Multiview frames are provided, corresponding to a robot policy rollout in the world model.  
Each frame consists of a fixed left view (upper right view), a fixed right view (upper left view), and a wrist view (bottom left).

\vspace{0.5em}
\textbf{Evaluation Method:}

\vspace{0.3em}
1. Use ALL three views (fixed left, fixed right, and wrist) for evaluation.\\
All views contribute equally to determining task success and robot behavior.

\vspace{0.5em}
\textbf{Provide:}

\vspace{0.3em}
1. A score (0 to \texttt{\{max\_score\}}) according to the rubric above\\
2. A brief explanation of your reasoning \\

\bottomrule
\end{tabular}
\caption{Prompt template used when the wrist view is used as a full evaluation signal.}
\label{tab:vlm_prompt_with_wrist}
\end{table*}

\subsection{Ablation on \ourmethod{} Evaluation Strategy}
\label{app:ablations}

We ablate the two design choices of \ourmethod's VLM-based evaluation 
(Section~\ref{app:vlm_rubric}; Table~\ref{tab:ablation_spearman_worldarena}). 
Replacing the rubric with a binary success-rate baseline discards the 
fine-grained scoring of task progress and world-model errors. Including 
the wrist view in success judgments---rather than restricting it to 
world-model failure detection---incurs a larger drop, indicating that 
the wrist view's higher artifact density biases the judge when used 
for success scoring.

\begin{table}[h]
\centering
\small
\setlength{\tabcolsep}{8pt}
\caption{Ablation on rank correlation with the RoboArena leaderboard 
(Spearman $\rho$). Numbers in parentheses indicate the drop relative 
to the full \ourmethod score.}
\label{tab:ablation_spearman_worldarena}
\begin{tabular}{lc}
\toprule
\textbf{Method / Ablation} & \textbf{Spearman $\rho$} \\
\midrule
\ourmethod score & 0.970 \;(\phantom{-}0.000) \\
\quad w/ wrist-view success judgments & 0.862 \;($-$0.108) \\
Success-rate baseline & 0.922 \;($-$0.048) \\
\bottomrule
\end{tabular}
\end{table}

\subsection{Consistent Trend across VLM Judges}
\label{app:vlm-judge-consistency}

\begin{figure}[t]
    \centering
    \includegraphics[width=\linewidth]{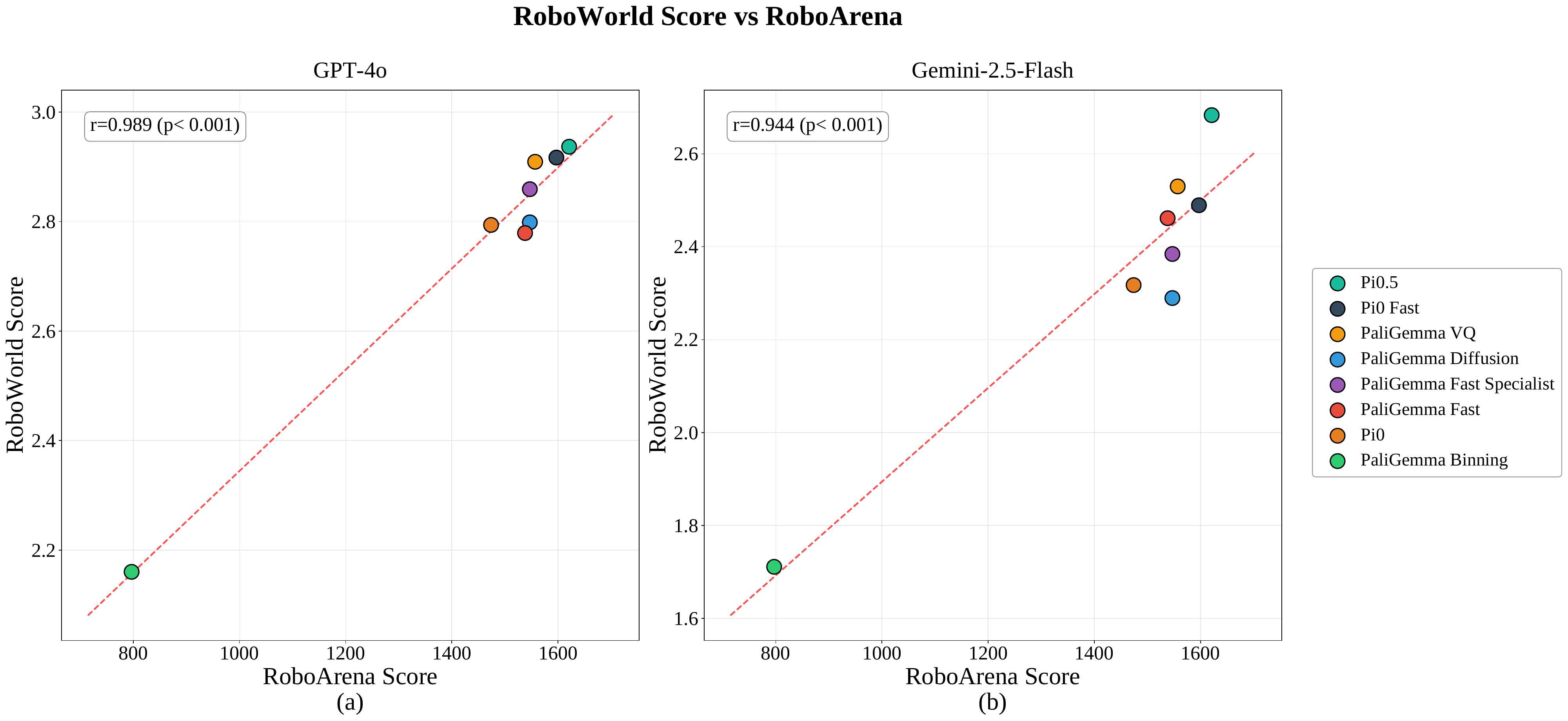}
    \caption{Pearson correlation between RoboArena score and \textbf{RoboWorld Score} under two different VLM judges: GPT-4o (a) and Gemini-2.5-Flash (b). Both judges show a strong linear relationship with the RoboArena leaderboard.}
    \label{fig:roboworld_score_pearson}
\end{figure}

\begin{figure}[t]
    \centering
    \includegraphics[width=\linewidth]{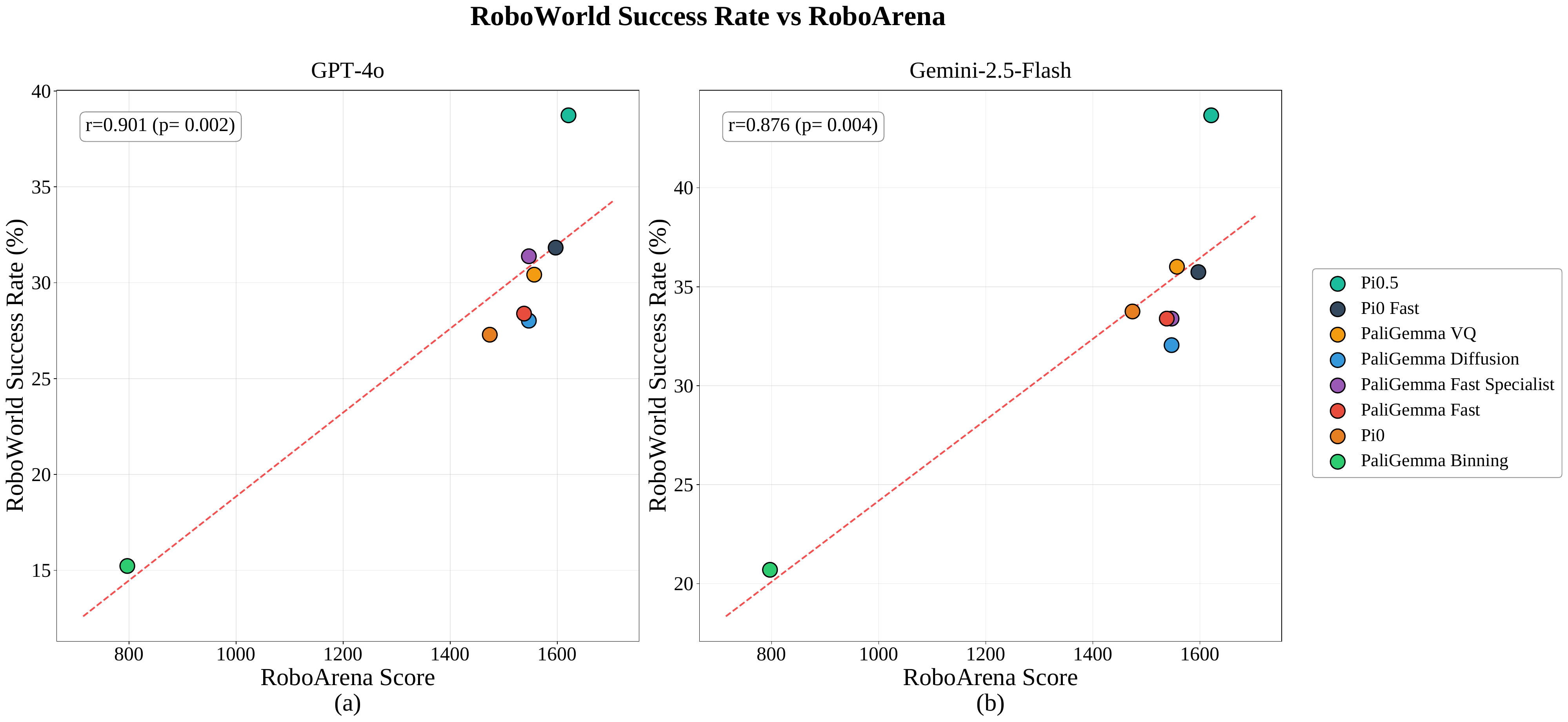}
    \caption{Pearson correlation between RoboArena score and \textbf{RoboWorld Success Rate} under two different VLM judges: GPT-4o (a) and Gemini-2.5-Flash (b). The success-rate signal also tracks the RoboArena ranking consistently across judges.}
    \label{fig:roboworld_sr_pearson}
\end{figure}

To evaluate whether RoboWorld metrics depend on a specific VLM judge, we additionally use \textit{Gemini-2.5-Flash} as an alternative judge. Figures~\ref{fig:roboworld_score_pearson} and~\ref{fig:roboworld_sr_pearson} compare the correlation between the RoboArena leaderboard score and RoboWorld metrics under two judges: GPT-4o and Gemini-2.5-Flash.

Across both judges, RoboWorld Score shows a strong and statistically significant linear relationship with the RoboArena score (Pearson $r{=}0.989$, $p{<}0.001$ for GPT-4o; $r{=}0.944$, $p{<}0.001$ for Gemini-2.5-Flash). The same trend holds for RoboWorld Success Rate, where both judges yield high Pearson correlations ($r{=}0.901$, $p{=}0.002$ for GPT-4o; $r{=}0.876$, $p{=}0.004$ for Gemini-2.5-Flash). The relative ordering of the eight policies is largely preserved across judges, and the fitted slopes are visually consistent. These results indicate that RoboWorld metrics are not tied to a particular VLM judge, but instead capture policy-level capability that aligns with closed-loop real-world evaluation.

\section{Implementation Details}
\label{app:implementation}

This section provides implementation details for video world model training
and interactive inference.

\subsection{Training}
\label{app:training}

Detailed training configurations are provided in
Table~\ref{table:training_details1} and Table~\ref{table:training_details2}.

\begin{table}[t]
\centering
\caption{Training details for Diffusion Forcing.}
\label{table:training_details1}
\small
\setlength{\tabcolsep}{6pt}
\begin{tabular}{ll}
\toprule
\textbf{Configuration} & \textbf{Setting} \\
\midrule

\multicolumn{2}{l}{\textit{Training data}} \\
\hspace{3mm}Input resolution & $368 \times 640$ \\
\hspace{3mm}Sequence length & 45 frames \\

\midrule
\multicolumn{2}{l}{\textit{Optimization}} \\
\hspace{3mm}Optimizer & AdamW \\
\hspace{3mm}Learning rate & $1\times10^{-5}$ \\
\hspace{3mm}Betas & $(0.9,\,0.99)$ \\
\hspace{3mm}$\epsilon$ & $1\times10^{-8}$ \\
\hspace{3mm}LR schedule & Constant with 1k-step warmup \\
\hspace{3mm}Gradient clipping & 1.0 \\
\hspace{3mm}Batch size & 8 \\
\hspace{3mm}Training steps & 160k \\
\hspace{3mm}Precision & Mixed precision (FP16) \\
\hspace{3mm}EMA decay & 0.9999 \\

\midrule
\multicolumn{2}{l}{\textit{Diffusion objective}} \\
\hspace{3mm}Training formulation & Rectified flow \\
\hspace{3mm}Prediction target & Velocity ($v$) \\
\hspace{3mm}Timestep schedule & Cosine (shift $=0.125$) \\
\hspace{3mm}Loss weighting & Sigmoid (bias $=-1.0$) \\

\bottomrule
\end{tabular}
\end{table}

\begin{table}[t]
\centering
\caption{Training details for \ouralgorithm.}
\label{table:training_details2}
\small
\setlength{\tabcolsep}{6pt}
\begin{tabular}{ll}
\toprule
\textbf{Configuration} & \textbf{Setting} \\
\midrule

\multicolumn{2}{l}{\textit{Training data}} \\
\hspace{3mm}Input resolution & $368 \times 640$ \\
\hspace{3mm}Sequence length & 45 frames \\

\midrule
\multicolumn{2}{l}{\textit{Optimization}} \\
\hspace{3mm}Optimizer & AdamW \\
\hspace{3mm}Learning rate & $1\times10^{-5}$ \\
\hspace{3mm}Betas & $(0.9,\,0.99)$ \\
\hspace{3mm}$\epsilon$ & $1\times10^{-8}$ \\
\hspace{3mm}Weight decay & $1\times10^{-3}$ \\
\hspace{3mm}LR schedule & Constant with 1k-step warmup \\
\hspace{3mm}Gradient clipping & 1.0 \\
\hspace{3mm}Batch size & 8 \\
\hspace{3mm}Training steps & 40k \\
\hspace{3mm}Precision & Mixed precision (FP16) \\
\hspace{3mm}EMA decay & 0.9999 \\

\midrule
\multicolumn{2}{l}{\textit{Diffusion objective}} \\
\hspace{3mm}Training formulation & Rectified flow \\
\hspace{3mm}Prediction target & Velocity ($v$) \\
\hspace{3mm}Timestep schedule & Cosine (shift $=0.125$) \\
\hspace{3mm}Loss weighting & Sigmoid (bias $=-1.0$) \\

\bottomrule
\end{tabular}
\end{table}

\subsection{Inference and Rollout Protocol}
All inference is performed autoregressively in the latent space of the
video VAE. The VAE compresses video temporally by a factor of $4\times$,
such that each latent token corresponds to four video frames. During
generation, latent tokens are produced causally, conditioned on previous
observations and the corresponding action sequence. Action conditioning is
provided at the token level by flattening the four 7-DoF robot actions
corresponding to each latent token.

\paragraph{Open-Loop Inference.}

For open-loop evaluation, the model receives the initial ground-truth
observation and the full ground-truth action sequence of the episode. The
model then generates all future observations autoregressively without
additional visual feedback. We evaluate the generated non-context frames
using PSNR, SSIM, LPIPS, FID, and FVD.

\paragraph{Closed-Loop Inference.}

In closed-loop inference, the world model serves as a neural simulator for
robot policy evaluation. Each rollout is initialized from the first
observation of a real episode. At every step, the latest generated
observation is provided to the policy, which predicts the next action
chunk. Conditioned on this action, the world model generates the next
latent token, which is decoded into video frames. The newest generated
observation is then fed back to the policy, forming a closed-loop
interaction cycle.

Final policy performance is evaluated from the generated videos using the
VLM-based scoring procedure described in Appendix~\ref{app:vlm-details}.

\section{Additional Results on Action-Conditioned Video Generation}
\label{app:additional-action-conditioned-generation}

Table~\ref{tab:droid_roboarena_301f_metrics_steps_avg} reports long-horizon
action-conditioned video generation results on RoboArena. Models generate
301-frame rollouts conditioned on multi-view observations and action sequences,
and metrics are averaged across all views. As shown in
Table~\ref{tab:droid_roboarena_301f_metrics_steps_avg}, Step Forcing achieves
the best perceptual quality among few-step methods, obtaining the lowest LPIPS
and FID with only 4 denoising steps. Notably, despite using substantially fewer
denoising steps, Step Forcing remains competitive with baselines that use 32 or
50 denoising steps, indicating that the proposed training objective enables
efficient long-horizon generation without requiring expensive sampling.

Table~\ref{tab:droid_roboarena_301f_metrics_steps_avg} also shows that IRASim
performs reasonably in short-horizon reproduction but degrades severely in our
301-frame rollout setting, especially in later frames, resulting in substantially
worse perceptual and temporal metrics. This suggests that short-horizon video
quality does not necessarily translate to stable long-horizon action-conditioned
rollout generation. PersistWorld improves over Ctrl-World on several pixel-level
metrics by applying post-training reinforcement learning on top of the
Ctrl-World framework. Since Step Forcing is orthogonal to such post-training
refinement, incorporating a similar post-training stage could further improve
our model.

\begin{table*}[tb!]
\centering
\small
\caption{Long-horizon video generation conditioned on multi-view frames and action sequences from RoboArena. Metrics are averaged over all views.}
\label{tab:droid_roboarena_301f_metrics_steps_avg}
\begin{tabular}{l c ccc ccc}
\toprule
Method & \makecell[c]{Denoising\\steps} &
\multicolumn{3}{c}{Pixel-level metrics} &
\multicolumn{3}{c}{Perceptual metrics} \\
\cmidrule(lr){3-5}\cmidrule(lr){6-8}
& & PSNR $\uparrow$ & MSE $\downarrow$ & SSIM $\uparrow$ & LPIPS $\downarrow$ & FID $\downarrow$ & FVD $\downarrow$ \\
\midrule
IRASim~\citep{zhu2024irasim} & 50 & 13.34 & 0.0516 & 0.518 & 0.469 & 119.67 & 1607 \\
Ctrl-World~\citep{guo2025ctrl}    & 50 & 16.56 & 0.0242 & 0.704 & 0.321 & 34.80 & 190.20 \\
PersistWorld~\citep{bardhan2026persistent} & 50 & 17.58 & 0.0194 & 0.728 & 0.305 & 34.70 & 197.60 \\
Diffusion Forcing~\citep{chen2024diffusion}                            & 50 & 15.27 & 0.0329 & 0.653 & 0.343 & 31.81 & 186.54 \\
\midrule
Ctrl-World~\citep{guo2025ctrl}    & 32 & 16.61 & 0.0237 & 0.706 & 0.322 & 35.38 & 190.80 \\
PersistWorld~\citep{bardhan2026persistent} & 32 & 17.56 & 0.0196 & 0.729 & 0.307 & 35.32 & 208.20 \\
Diffusion Forcing~\citep{chen2024diffusion}                            & 32 & 15.35 & 0.0323 & 0.661 & 0.338 & 30.07 & \textbf{178.96} \\
\midrule
Ctrl-World~\citep{guo2025ctrl}    & 16 & 16.66 & 0.0232 & 0.712 & 0.324 & 38.16 & 229.80 \\
PersistWorld~\citep{bardhan2026persistent} & 16 & \textbf{17.66} & \textbf{0.0191} & \textbf{0.732} & 0.309 & 37.44 & 249.60 \\
Diffusion Forcing~\citep{chen2024diffusion}                            & 16 & 15.52 & 0.0313 & 0.672 & 0.331 & 29.16 & 190.21 \\
\midrule
Ctrl-World~\citep{guo2025ctrl}    & 8  & 16.61 & 0.0232 & 0.712 & 0.333 & 43.83 & 317.20 \\
PersistWorld~\citep{bardhan2026persistent} & 8  & 17.62 & 0.0193 & 0.731 & 0.314 & 40.64 & 306.30 \\
Diffusion Forcing~\citep{chen2024diffusion}                            & 8  & 15.69 & 0.0302 & 0.678 & 0.327 & 29.98 & 210.54 \\
\midrule
  Ctrl-World~\citep{guo2025ctrl}    & 4  & 15.39 & 0.0303 & 0.664 & 0.392 & 69.17 & 529.70 \\
PersistWorld~\citep{bardhan2026persistent} & 4  & 17.01 & 0.0222 & 0.709 & 0.341 & 51.13 & 403.00 \\
Diffusion Forcing~\citep{chen2024diffusion}                            & 4  & 15.81 & 0.0292 & 0.683 & 0.327 & 33.18 & 242.00 \\
\midrule
\textbf{Step Forcing (Post-Train)}      & 4  & 16.79 & 0.0230 & 0.715 & \textbf{0.293} & 28.54 & 203.83 \\
\textbf{Step Forcing (Pre-train)}      & 4  & 16.92 & 0.0220 & 0.713 & 0.295 & \textbf{28.28} & 197.3 \\
\textbf{Step Forcing (Pre-train 240k step)}      & 4  & 16.92 & 0.0220 & 0.714 & 0.295 & 28.29 & 192.00 \\
\bottomrule
\end{tabular}
\end{table*}

\section{Closed-Loop Evaluation Details and Qualitative Analysis}
\label{app:closed_loop_qualitative}

\subsection{\ourmethod Implementation Details}

\ourmethod evaluates a policy entirely inside the world model: the policy never
interacts with a real robot, and all observations it receives are generated
frames.

\paragraph{Episode specification.}
Each episode in \ourmethod is defined by (i) an initial multi-view observation
(a $2{\times}2$ collage of two exterior views and a wrist view), (ii) a language
instruction, and (iii) the initial robot state recorded at the first timestep of
the original session. Background-augmented episodes share the instruction and
initial robot state of their source session and differ only in the initial
image.

\paragraph{Policy interface.}
At every step, the evaluation harness crops the exterior and wrist views from
the latest generated collage, resizes them to the policy's input resolution, and
passes them to the policy together with the current proprioceptive state and the
episode's language instruction. The policy returns an action chunk in its native
action space, consisting of joint-velocity commands and a gripper command. The
language instruction is consumed only by the policy; the world model is
conditioned on visual history and robot state/action information, and never
observes text.

\paragraph{State and action conversion.}
The policies evaluated in \ourmethod output joint-space actions, whereas the
world model is conditioned on end-effector state. Therefore,
the evaluation harness maintains the robot state and converts policy outputs
into the conditioning format required by the world model. Specifically, the
joint state is initialized from the episode's initial robot state and updated by
integrating the policy's predicted joint-velocity commands. The updated joint
state is used to construct the proprioceptive observation provided back to the
policy. In parallel, we compute the corresponding end-effector poses through
forward kinematics and use them, together with the gripper command, as the
state/action conditioning for the world model.

This bookkeeping ensures that the policy input and the world-model conditioning
are derived from the same kinematically consistent robot trajectory, while
preserving their respective input representations.

\subsection{Qualitative Results}

\begin{figure}[t]
  \centering
  \includegraphics[width=\linewidth]{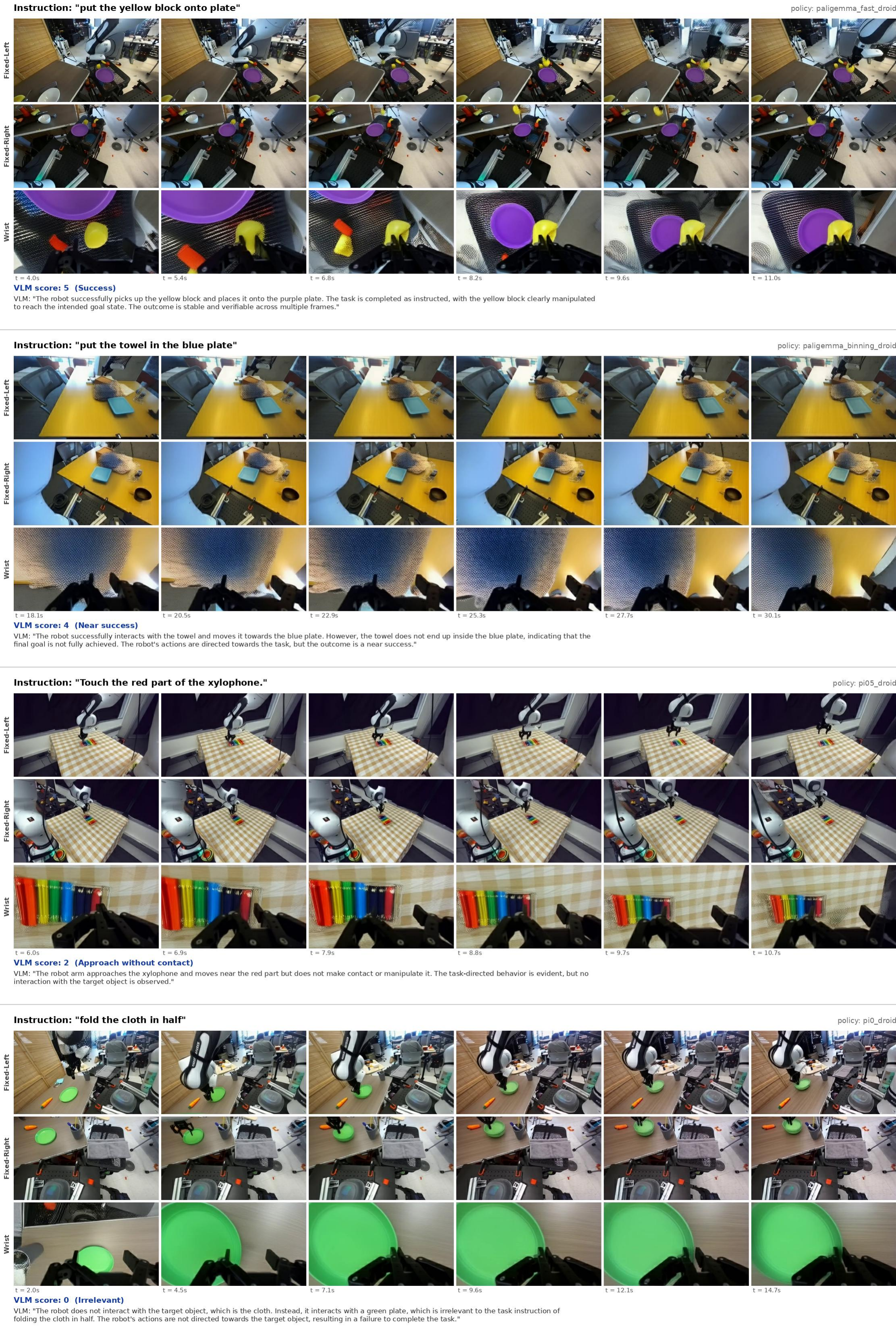}
  \caption{
  World-model rollouts for each level of the task-progress-aware rubric. From top to bottom: success (5), near success (4), approach without contact (2), and irrelevant behavior (0), each shown with the VLM's verbatim judgment.
  }

  \label{fig:roboworld_examples}
\end{figure}

We provide qualitative examples of world-model rollouts together with the
corresponding VLM annotations for each case. As shown in
Figure~\ref{fig:roboworld_examples}, these examples make it easier to interpret
how the evaluator scores different rollout behaviors, including success, near
success, partial progress, and irrelevant behavior. We further analyze both
world-model failures and VLM evaluation failures in the following examples.

\subsection{Failure Case Analysis}

\begin{figure}[t]
  \centering
  \includegraphics[width=\linewidth]{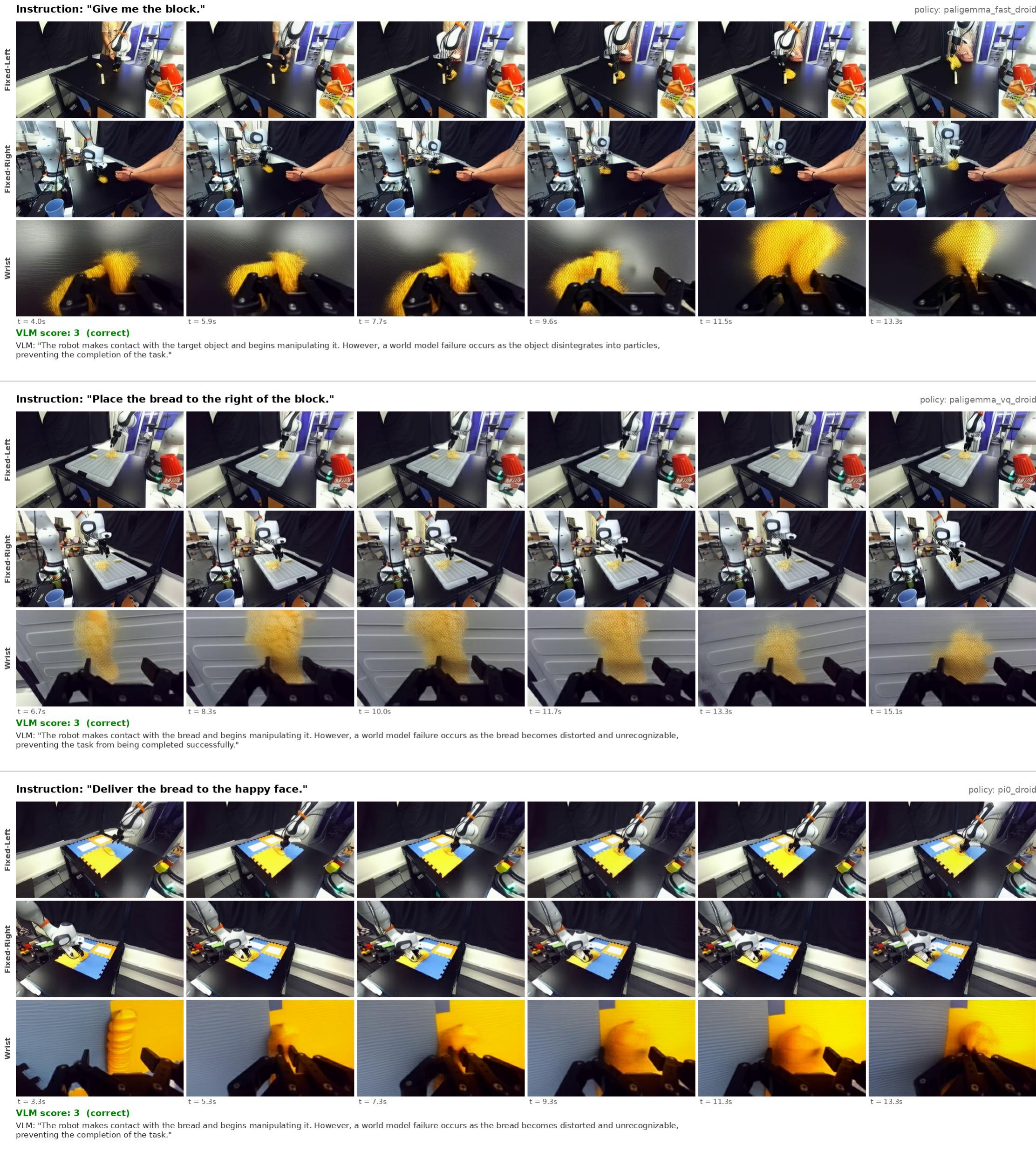}
  \caption{
  Qualitative examples of world-model failures correctly detected by the VLM evaluator. After the robot makes contact, the manipulated object disintegrates or morphs into unrecognizable artifacts; the task-progress-aware rubric assigns its dedicated score (3) to these cases.
  }

  \label{fig:wm_failures}
\end{figure}

\paragraph{World Model Failure Cases}
Figure~\ref{fig:wm_failures} shows qualitative examples of world-model failures
that are correctly identified by the VLM evaluator. We observe that such failures
occur primarily during object interaction. Before contact, the generated scene is
typically stable, but after the robot begins manipulating the object, the object
may disintegrate, morph into unrealistic shapes, or become visually inconsistent.
This suggests that contact-rich object dynamics remain a key limitation of the
current world model. We expect these failures to be mitigated by scaling to
larger-capacity world models and improving interaction modeling.

\begin{figure}[t]
  \centering
  \includegraphics[width=\linewidth]{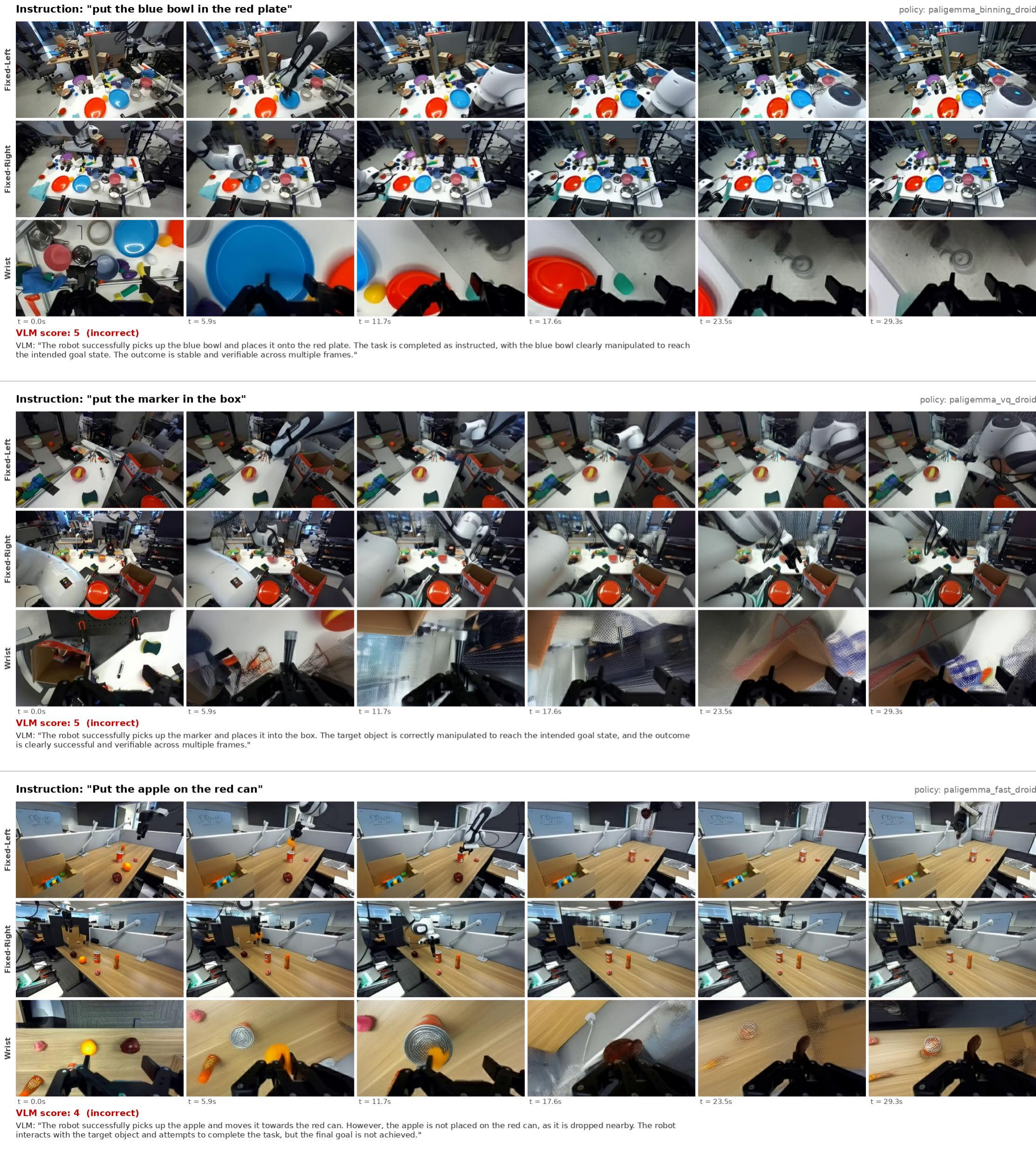}
  \caption{
  Qualitative examples of VLM evaluation failures. GPT-4o assigns success or near-success scores (5, 5, 4) to rollouts in which the scene remains essentially unchanged, hallucinating task completion that the video contradicts.
  }
  \label{fig:vlm_failures}
\end{figure}

\paragraph{VLM Failure Cases}
Figure~\ref{fig:vlm_failures} shows qualitative examples of VLM evaluation
failures. While the VLM evaluator generally captures the relative trends across
policies, we find that it tends to assign more lenient scores than human
judgment. In a manual comparison over approximately 100 sampled rollouts, the
average VLM score was about one point higher than the corresponding human score.
The examples in Figure~\ref{fig:vlm_failures} illustrate this behavior: GPT-4o
occasionally assigns success or near-success scores to rollouts where the scene
remains essentially unchanged, hallucinating task completion despite visual
evidence to the contrary.

\end{document}